\documentclass{article}

\usepackage[main, final]{neurips_2026}

\usepackage[utf8]{inputenc} 
\usepackage[T1]{fontenc}    
\usepackage{url}            
\usepackage{booktabs}       
\usepackage{amsfonts}       
\usepackage{nicefrac}       
\usepackage{xcolor}         
\usepackage{amsmath}
\usepackage{graphicx}
\usepackage{multirow}
\usepackage{algorithm}
\usepackage{algpseudocode}
\usepackage{hyperref}       
\usepackage[table]{xcolor}
\definecolor{PGSBblue}{RGB}{232,242,255}
\newcommand{\PGSBcell}[1]{\cellcolor{PGSBblue}#1}
\newcommand{\PGSBavg}[1]{\cellcolor{PGSBblue}\textbf{#1}}
\usepackage{array}
\newcolumntype{C}[1]{>{\centering\arraybackslash}p{#1}}
\newcolumntype{L}[1]{>{\raggedright\arraybackslash}p{#1}}
\title{PGSB: Pretrained-Guided Shared Basis for LoRA Model Merging}

\author{%
  Muqing Liu$^{1, \dagger}$,\quad 
    Chongjie Si$^{2, \dagger}$,\quad 
    Zhuoya Liu$^{1}$,\quad 
    Yuheng Jia$^{3,4}$\thanks{Corresponding author. \\ \protect \indent $^{\dagger}$Equal contribution.}\\
$^{1}$Chien-Shiung Wu College, Southeast University 
$^{2}$MoE Key Lab of Artificial Intelligence, AI\\ Institute, School of Computer Science, Shanghai Jiao Tong University $^{3}$School of Computer\\ Science and Engineering, Southeast University
$^{4}$Key Laboratory of New Generation Artificial\\ Intelligence Technology and Its Interdisciplinary Applications (Southeast University), Ministry \\of Education, China\\
{\tt \small liumq23@seu.edu.cn \ chongjiesi@sjtu.edu.cn \  zhuoyaliu@seu.edu.cn \  yhjia@seu.edu.cn} \\ \\
}

\begin{document}

\maketitle
\begin{abstract}

Large pretrained models have achieved remarkable success across diverse domains, yet adapting them to multiple downstream tasks remains challenging: full fine-tuning is costly, while maintaining separate task-specific models is storage-inefficient and fails to produce a unified multi-task model.
Low-Rank Adaptation (LoRA) offers an efficient alternative for Model Merging by representing each task adaptation as a low-rank update over frozen pretrained weights.
However, existing LoRA merging methods often treat task-specific updates as directly comparable objects and merge them in their original low-rank parameter spaces.
This overlooks two key factors: which update directions provide reliable shared support across tasks, and how these directions should be re-parameterized with respect to the pretrained weights before merging.
To address this limitation, we propose Pretrained-Guided Shared Basis (PGSB), a two-stage framework for LoRA model merging.
PGSB first identifies shared directions among task-specific LoRA updates, and then re-parameterizes these directions using their interaction with the pretrained weights, producing a more suitable basis for merging.
Extensive experiments demonstrate that PGSB consistently outperforms state-of-the-art baselines across multiple benchmarks.
Codes are available at \url{https://github.com/Kitschnada/Pretrained-Basis}.
\end{abstract}

\section{Introduction}
Large pretrained models have demonstrated strong performance across diverse tasks~\cite{devlin2018bert,kirillov2023segment,liu2019roberta}, making them powerful foundations for downstream adaptation.
As independently adapted models become increasingly available through model hubs such as Hugging Face~\cite{huggingface}, model merging has emerged as an attractive paradigm for constructing a unified multi-task model by combining task-specific models without requiring access to their original training data~\cite{ilharco2022editing,matena2022merging,si2025nan,wortsman2022model,yadav2024ties}.
However, most existing merging methods are designed for fully fine-tuned models~\cite{ilharco2022editing,yadav2024ties,yang2024adamerging,yu2024language}, which presupposes that each task model is obtained by updating all pretrained parameters.
This requirement becomes increasingly costly in computation, memory, and storage as model scale grows~\cite{ma2024segment,qiu2020pre,raffel2020exploring}.
In contrast, parameter-efficient fine-tuning (PEFT) methods, such as Low-Rank Adaptation (LoRA)~\cite{tang2023parameter}, learns compact task-specific adaptations over frozen pretrained weights with extremely low costs while retaining competitive performance~\cite{he2021towards,houlsby2019parameter,hu2021lora,liu2026flex,si2024flora,zhang2023adalora}.
Therefore, merging PEFT-based adaptations has attracted growing research attention as a scalable route toward unified multi-task models.

Despite this promise, directly extending existing merging methods to PEFT-based adaptations remains non-trivial.
In LoRA, each task adaptation is represented by a pair of low-rank matrices rather than a full weight update, making its merging geometry fundamentally different from that of fully fine-tuned models.
Recent methods such as KnOTS and Core Space~\cite{stoica2024knots,panariello2026core} address this issue by projecting LoRA updates into a shared subspace before merging.
Although these methods have achieved promising results, they do not fully resolve the core difficulty of LoRA merging.
A more principled solution must answer two challenging questions: how to distill a unified set of directions from diverse task-specific LoRA updates so that it can carry information shared across multiple tasks, and how to represent these directions in a way that remains compatible with the scale and geometry of the original pretrained weights.

To answer these questions, we propose a novel framework named \textbf{Pretrained-Guided Shared Basis} (PGSB) for LoRA model merging.
PGSB consists of two stages: \emph{Task-shared Subspace Extraction} (TSE) and \emph{Pretrained-coupled Coordinate Refinement} (PCR), which accordingly determines the shared task directions and their pretrained-compatible merge coordinates.
In the first stage, PGSB concatenates task-specific LoRA updates and performs singular value decomposition to extract a shared subspace, yielding a unified multi-task basis that captures common task-relevant directions.
In the second stage, PGSB refines this basis using the geometric structure of the pretrained weights, producing a pretrained-compatible multi-task basis that reduces conflicts with the original weight space.
By merging LoRA updates in this refined basis, PGSB better preserves task-specific knowledge while maintaining compatibility with the pretrained model, leading to more reliable multi-task adaptation.
Extensive experiments on vision and language benchmarks show that PGSB consistently outperforms strong baselines and achieves state-of-the-art results.
Our contributions are as follows:

\begin{itemize}
    \item We show that effective LoRA merging requires both constructing a shared multi-task basis and ensuring its compatibility with the pretrained weight geometry.

    \item We propose PGSB, a two-stage LoRA merging framework that first extracts a task-shared subspace from concatenated LoRA updates and then refines the resulting basis using the geometric structure of pretrained weights.

    \item Extensive experiments on vision and language benchmarks demonstrate that PGSB consistently outperforms strong model merging baselines, validating the effectiveness of pretrained-guided basis refinement for scalable multi-task adaptation.
\end{itemize}

\section{Related Work}

\subsection{Model Merging}

With the proliferation of task-specific models, integrating them into a single model capable of handling multiple tasks without additional training has become a central problem in model merging~\cite{matena2021merging}.
A representative approach is task arithmetic~\cite{ilharco2022editing}, which constructs task vectors by subtracting base-model parameters from fine-tuned models and combines them through linear aggregation.
However, it often suffers from performance degradation due to interference among task updates.
To mitigate this issue, subsequent methods introduce more structured merging strategies~\cite{Akiba2025evolutionary,si2025nan,jin2022dataless,panariello2025modular,marouf2023weighted}.
TIES~\cite{yadav2024ties} resolves sign conflicts and filters updates by magnitude, while DARE~\citep{yu2024language} reduces redundancy through random dropping and rescaling.
More recent methods, such as TSV~\citep{tsv} and Iso-C~\citep{marczak2025task}, further exploit the matrix structure of task updates, highlighting the importance of geometric alignment in effective merging.

Despite these advances, methods designed for fully fine-tuned models do not transfer well to LoRA-based adaptations~\cite{tang2023parameter}.
In LoRA, task adaptations are represented as low-rank factors, leading to a fundamentally different merging geometry.
KnOTS~\cite{stoica2024knots} addresses this by merging in a shared subspace, but incurs high computational cost due to SVD on concatenated full-size matrices.
Core Space~\cite{panariello2026core} improves efficiency via a shared low-rank space, yet relies on a fixed linear representation and insufficiently models interactions between task updates and pretrained weights.
Thus, achieving efficient and reliable LoRA merging while preserving task information remains an open challenge.
Detailed formulations of KnOTS and Core Space are deferred to Appendix~\ref{app:prior_alignment_methods}.

\subsection{Parameter-efficient Fine Tuning (PEFT)}
Large pretrained models have been applied in various downstream tasks, enabling significant improvement in generalization.
Due to the huge cost fully fine-tuning brings to computation and storage, Parameter-efficient fine-tuning (PEFT) methods have been proposed, which uses only a small subset of parameters without reducing the performances of fully fine-tuning.
PEFT techniques include strategies such as Adapter~\cite{he2021towards,pfeiffer2020adapterfusion,houlsby2019parameter}, Prompt Tuning~\cite{li2021prefix,shi2023dept,wang2023non,fischer2024prompt}, and Low-Rank Adaptation (LoRA)~\cite{hu2021lora,zhang2022adaptive,si2024flora,si2024see,liu2026flex}.
Among these, LoRA~\cite{hu2021lora} is widely adopted for its simplicity and effectiveness.
It freezes the pretrained weights and learns task-specific low-rank updates for linear layers, enabling efficient adaptation with few trainable parameters.
Since the learned updates can be merged into the original weights during inference, LoRA preserves the model architecture without introducing additional latency.

\section{Method}

\subsection{Preliminary}

\paragraph{Low-Rank Adaptation.} 
For a pretrained weight matrix $\mathbf{W} \in \mathbb{R}^{d_{\text{out}} \times d_{\text{in}}}$, LoRA freezes $\mathbf{W}$ and introduces a trainable low-rank update $\Delta \mathbf{W}$:
\begin{equation}
    \mathbf{W}' = \mathbf{W} + \Delta \mathbf{W}, 
    \quad \Delta \mathbf{W} = \mathbf{B}\mathbf{A},
\end{equation}
where $\mathbf{A} \in \mathbb{R}^{r \times d_{\text{in}}}$ and $\mathbf{B} \in \mathbb{R}^{d_{\text{out}} \times r}$ are the only trainable parameters.
The rank $r \ll \min(d_{\text{out}}, d_{\text{in}})$ is chosen to be small, so that each task adaptation can be learned with substantially fewer parameters than full fine-tuning.

\paragraph{Model Merging.}
Given a set of task-specific models $\{\mathbf{W}_t\}_{t=1}^T$ fine-tuned from the same pretrained model $\mathbf{W}_0$, model merging aims to construct a single model that integrates knowledge from multiple tasks without additional training. 
A common formulation is to first define the task vector of each model as the difference from the pretrained weights~\cite{ilharco2022editing}, i.e., $\tau_t = \mathbf{W}_t - \mathbf{W}_0$. 
Then, a merging operator $\mathcal{M}$ is applied to obtain a merged task vector:
\begin{equation}
    \tau^\star = \mathcal{M}(\tau_1,\tau_2,\dots,\tau_T).
\end{equation}
Finally, the merged model is reconstructed as
\begin{equation}
    \mathbf{W}_{\mathrm{merge}} = \mathbf{W}_0 + \tau^\star.
\end{equation}

\subsection{Overview}
\begin{figure}
    \centering
    \includegraphics[width=\linewidth]{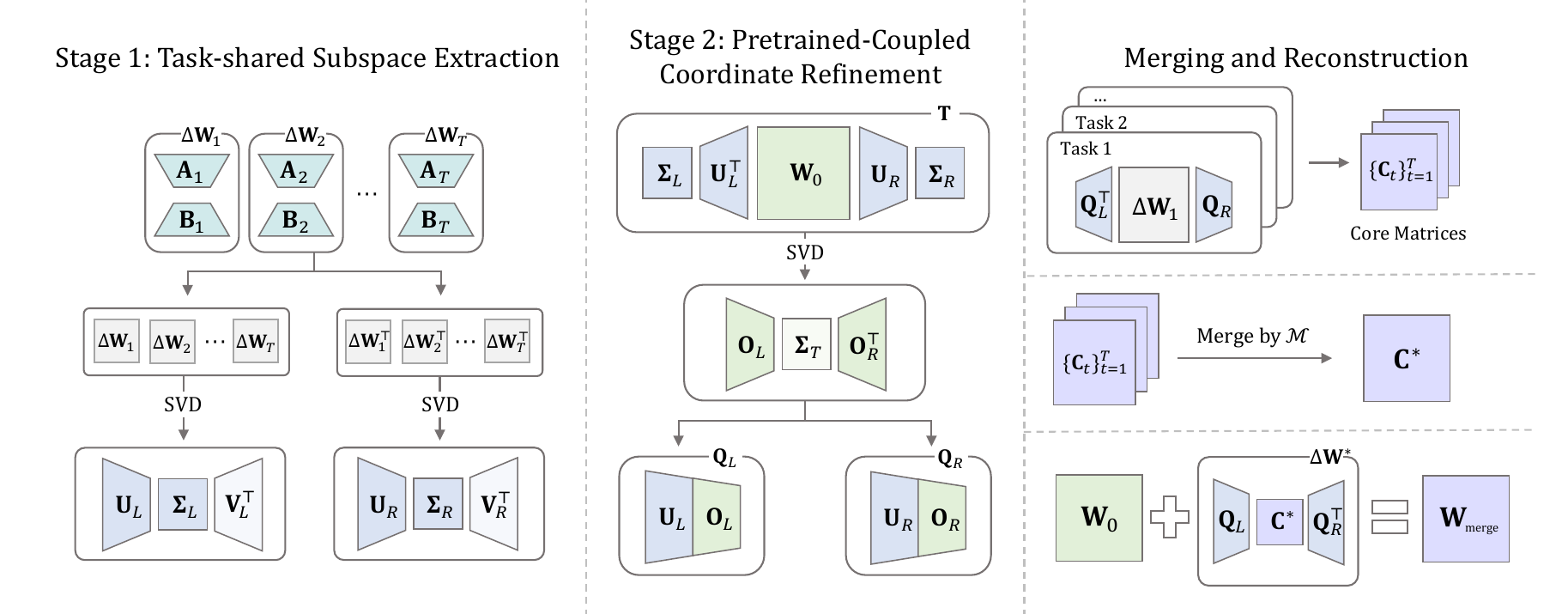}
    \caption{Overview of PGSB.Given task-specific LoRA updates, PGSB first performs Task-shared Subspace Extraction by decomposing the concatenated updates and their transposed counterparts, producing shared support matrices. PGSB then performs Pretrained-Coupled Coordinate Refinement, where the extracted supports are coupled with the pretrained weight geometry to obtain pretrained-compatible multi-task bases. Finally, task updates are projected into the refined bases, merged by a downstream merge operator $\mathcal{M}$, and reconstructed back to the original parameter space to obtain the final merged model.}
    \label{fig:main}
\end{figure}

PGSB separates LoRA merging into two stages: task-shared subspace extraction
(TSE) and pretrained-coupled coordinate refinement (PCR). TSE extracting a task-shared subspace that provides a unified multi-task basis, and PCR refines this basis with the pretrained weight geometry to obtain pretrained-compatible merge coordinate.
The following sections describe these two stages in detail.
The framework of PGSB is shown in Fig. \ref{fig:main}.

\subsection{Task-shared Subspace Extraction}

PGSB first constructs a shared pair of support matrices for representing multiple LoRA updates.
Since each update $\Delta \mathbf{W}_t$ is a matrix, its action involves two coupled spaces: the column space and the row space.
Therefore, a single shared basis is insufficient; we need to extract common structures from both sides of the updates.
We refer to the resulting bases as the left and right support matrices.

Specifically, PGSB first concatenates all task-specific updates and applies singular value decomposition (SVD):
\begin{equation}
    [\Delta \mathbf{W}_1,\Delta \mathbf{W}_2,\dots,\Delta \mathbf{W}_T]
    =
    \mathbf{U}_L \mathbf{\Sigma}_L \mathbf{V}_L^\top,
\end{equation}
where $\mathbf{U}_L$ captures the shared column-space structure of task updates and is used as the left support matrix.
To extract the shared row-space structure, we equivalently apply SVD to the concatenated transposed updates:
\begin{equation}
    [\Delta \mathbf{W}_1^\top,\Delta \mathbf{W}_2^\top,\dots,\Delta \mathbf{W}_T^\top]
    =
    \mathbf{U}_R \mathbf{\Sigma}_R \mathbf{V}_R^\top.
\end{equation}
where $\mathbf{U}_R$ is used as the right support matrix.
The transposition is introduced because the row space of $\Delta \mathbf{W}_t$ becomes the column space of $\Delta \mathbf{W}_t^\top$, allowing both support matrices to be obtained through the same SVD-based construction.

\subsubsection{Efficient Low-Rank Support Construction}

Although the full-update formulation provides a direct definition of the task-shared support pair, explicitly factorizing concatenated full-size matrices is computationally expensive and undermines the efficiency advantage of LoRA.
Since each LoRA update is intrinsically low-rank: $\Delta \mathbf{W}_t=\mathbf{B}_t\mathbf{A}_t$, we have $\operatorname{rank}(\Delta \mathbf{W}_t)\leq r$.
Therefore, the shared support subspaces can be recovered from compact low-rank factors rather than from the full matrices.

\begin{table}[!ht]
\centering   
\renewcommand\arraystretch{0.9} 
\setlength{\tabcolsep}{7mm}
\caption{Asymptotic preprocessing complexity for constructing the merging space of one layer before applying a downstream merge operator. We assume square weight matrices of size $n \times n$, $T$ tasks, and LoRA rank $r$ with $Tr \ll n$.}
\resizebox{\textwidth}{!}{
    \begin{tabular}{lcc}
    \toprule
    Method & Construction object & Complexity \\
    \midrule
    Baseline & Directly merge $\Delta \mathbf{W}_t=\mathbf{B}_t\mathbf{A}_t$ & $\mathcal{O}(n^2Tr)$ \\
    KnOTS & SVD on $[\Delta \mathbf{W}_1,\dots,\Delta \mathbf{W}_T]$ & $\mathcal{O}(n^3T^2)$ \\
    Core Space & SVD on stacked low-rank factors & $\mathcal{O}(n^2Tr + T^3r^3)$ \\
    PGSB & Two full-space SVDs & $\mathcal{O}(n^3T^2 + n^2Tr + T^3r^3)$ \\
    PGSB (efficient) & Two low-rank SVDs & $\mathcal{O}(n^2Tr + T^3r^3)$ \\
    \bottomrule
    \end{tabular}
}
\label{tab:complexity}
\end{table}

For each task $t$, let
$\Delta \mathbf{W}_t = \mathbf{U}_t \mathbf{\Sigma}_t \mathbf{V}_t^\top$
be its compact SVD, where $\mathbf{U}_t$ and $\mathbf{V}_t$ contain only the nonzero singular directions.
We define
\begin{equation}
\mathbf{L} = [\mathbf{U}_1\mathbf{\Sigma}_1,\dots,\mathbf{U}_T\mathbf{\Sigma}_T],
\qquad
\mathbf{R} = [\mathbf{V}_1\mathbf{\Sigma}_1,\dots,\mathbf{V}_T\mathbf{\Sigma}_T].
\end{equation}
Since each block has at most $r$ columns, $\mathbf{L}$ and $\mathbf{R}$ have only $Tr$ columns, making the decomposition substantially cheaper than SVD on concatenated full updates.
We then perform SVD on these low-rank concatenations:
\begin{equation}
    \mathbf{L}=\bar{\mathbf{U}}_L\bar{\mathbf{\Sigma}}_L\widetilde{\mathbf{V}}_L^\top,
    \qquad
    \mathbf{R}=\bar{\mathbf{U}}_R\bar{\mathbf{\Sigma}}_R\widetilde{\mathbf{V}}_R^\top,
\end{equation}
and use $\bar{\mathbf{U}}_L$ and $\bar{\mathbf{U}}_R$ as the left and right support matrices.
This yields the same support matrices and singular values (i.e., $\mathbf{U}=\bar{\mathbf{U}}_L$, $\mathbf{\Sigma}_L= \begin{bmatrix}
    \bar{\mathbf{\Sigma}}_L & \mathbf{0}
\end{bmatrix}$, $\bar{\mathbf{U}}_R=\mathbf{U}_R$, and $\mathbf{\Sigma}_R= \begin{bmatrix}
    \bar{\mathbf{\Sigma}}_R & \mathbf{0}
\end{bmatrix}$) as the full-update formulation, while avoiding SVD on large concatenated matrices.
As shown in Table~\ref{tab:complexity}, this reduces the computational complexity while preserving the original task-shared support construction; the equivalence proof and detailed complexity derivation are provided in Appendices~\ref{app:support_construction} and~\ref{app:complexity}, respectively.

\subsection{Pretrained-Coupled Coordinate Refinement}

TSE identifies a shared support pair $\mathbf{U}_L$ and $\mathbf{U}_R$ for LoRA updates.
These support matrices specify a common subspace in which task updates can be represented, but they do not uniquely determine the coordinate system inside this subspace.
Indeed, the same subspace can admit many equivalent bases.
For any orthogonal matrices $\mathbf{R}_L$ and $\mathbf{R}_R$, $\mathbf{U}_L\mathbf{R}_L$ and $\mathbf{U}_R\mathbf{R}_R$ span exactly the same subspaces as
$\mathbf{U}_L$ and $\mathbf{U}_R$, and thus are equally valid support matrices, yet they produce different coordinate representations for the same update.
This distinction is crucial because practical merge rules, such as TIES~\cite{yadav2024ties}, are coordinate-sensitive: magnitude trimming, sign selection, and conflict resolution are performed entry by entry and therefore depend on the chosen basis.

Therefore, after obtaining the task-shared supports, we further seek orthogonal transformations that preserve the shared subspaces while selecting a more suitable coordinate system for merging.
Since every LoRA update is learned as an offset to the same pretrained weight $\mathbf{W}_0$, the merge coordinate should also respect how these shared directions are arranged in the pretrained model.
To measure this relation, we project the pretrained weight onto the task-shared supports:
\begin{equation}
    \mathbf{G} = \mathbf{U}_L^\top \mathbf{W}_0 \mathbf{U}_R.
\end{equation}
The matrix $\mathbf{G}$ describes how the pretrained weight is expressed under the task-shared coordinates.
In particular, its entries reflect the coupling between shared directions induced by $\mathbf{W}_0$.
Meanwhile, the singular values $\mathbf{\Sigma}_L$ and $\mathbf{\Sigma}_R$ quantify the aggregate importance of the corresponding shared directions across task updates themselves.
We therefore use them to rescale the pretrained coupling:
\begin{equation}
    \mathbf{T}
    =
    \mathbf{\Sigma}_L \mathbf{G} \mathbf{\Sigma}_R .
\end{equation}
Thus, $\mathbf{T}$ represents the pretrained-weight coupling after taking into account the relative importance of task-shared directions.
However, $\mathbf{T}$ is generally non-diagonal, which means that the current task-shared coordinates do not separate the dominant interactions well.
For example, if one row of $\mathbf{T}$ contains several large off-diagonal entries, then one shared direction is strongly coupled with multiple directions on the other side.
In this case, the effect of this direction is distributed across multiple coordinate entries, rather than being concentrated in a single dominant coordinate.
This is undesirable for coordinate-wise merge rules such as TIES, where magnitude trimming, sign selection, and conflict resolution are performed independently for each entry.
Such independent entry-wise decisions may break correlated interactions that should ideally be treated together.

To reduce this coordinate mixing, we compute the SVD of $\mathbf{T}$:
\begin{equation}
    \mathbf{T} = \mathbf{O}_L \mathbf{\Sigma}_T \mathbf{O}_R^\top.
\end{equation}
After this rotation, the dominant interactions are concentrated on diagonal coordinates, making each coordinate more clearly associated with one major pretrained-compatible interaction.

We combine the task-shared supports from TSE with the pretrained-coupled rotations:
\begin{equation}
    \mathbf{Q}_L = \mathbf{U}_L \mathbf{O}_L,
    \qquad
    \mathbf{Q}_R = \mathbf{U}_R \mathbf{O}_R.
\end{equation}
The resulting bases $\mathbf{Q}_L$ and $\mathbf{Q}_R$ preserve the original task-shared subspaces, while providing a coordinate system refined by the pretrained weight geometry.

\subsection{Merging and Reconstruction}

After obtaining the refined bases $\mathbf{Q}_L$ and $\mathbf{Q}_R$, PGSB first maps all task-specific LoRA updates into the same pretrained-compatible coordinate space.
This gives a set of task matrices $\{\mathbf{C}_t\}_{t=1}^{T}$:
\begin{equation}\label{eq:PGSB}
    \mathbf{C}_t
    =
    \mathbf{Q}_L^\top \Delta \mathbf{W}_t \mathbf{Q}_R,
\end{equation}
where each $\mathbf{C}_t$ represents the corresponding update under the shared task subspace and the refined merge coordinate.
Any merge rule $\mathcal{M}$ can then be applied in this coordinate space:
\begin{equation}
    \mathbf{C}^\star
    =
    \mathcal{M}(\mathbf{C}_1,\mathbf{C}_2,\dots,\mathbf{C}_T).
\end{equation}
Finally, the merged coordinate matrix is mapped back to the original parameter space:
\begin{equation}
    \Delta \mathbf{W}^\star
    =
    \mathbf{Q}_L \mathbf{C}^\star \mathbf{Q}_R^\top .
\end{equation}
Thus, PGSB performs merging in a basis that retains the task-shared subspace structure while adapting its coordinates to the pretrained weight geometry, and then reconstructs the merged LoRA update for deployment.
The specific algorithm of PGSB is deferred to Appendix~\ref{app:algo}.

\section{Experiment}

\subsection{Experimental Setting}

\paragraph{Models.}
We evaluate PGSB on both vision and language model merging benchmarks.
For vision experiments, we use CLIP~\cite{radford2021learning} with ViT-B/32 and ViT-L/14 backbones~\cite{dosovitskiy2020image} and merge LoRA adapters fine-tuned on eight image classification tasks.
For language experiments, we use Llama-3-8B~\cite{llama3modelcard} and merge LoRA adapters fine-tuned on six Natural Language Inference (NLI) tasks.
We also use Llama-3-8B~\cite{llama3modelcard} and merge LoRA adapters fine-tuned on eight Commonsense Reasoning tasks.
All LoRA checkpoints and evaluation protocols follow prior alignment-based LoRA merging studies~\cite{stoica2024knots,panariello2026core}.

\paragraph{Datasets.}
The vision benchmark covers diverse recognition scenarios, including texture classification, digit recognition, scene understanding, and remote sensing, with representative datasets such as DTD~\cite{dtd}, EuroSAT~\cite{eurosat}, MNIST~\cite{MNIST}, and SUN397~\cite{sun397}.
The language benchmark covers standard NLI, entailment-style tasks and reasoning tasks, including MNLI~\cite{williams2017broad},  SciTail~\cite{scitail}, and BoolQ~\cite{clark2019boolq}.

\paragraph{Baselines.}
We compare PGSB with representative merge spaces, including KnOTS~\cite{stoica2024knots} and Core Space~\cite{panariello2026core}.
We also report the \textbf{Baseline}, where each merge operator is directly applied to the reconstructed LoRA updates $\Delta\mathbf{W}_t=\mathbf{B}_t\mathbf{A}_t$ without any space transformation.
For all merge spaces, we use the same merge operators, including TA~\cite{ilharco2022editing}, TIES~\cite{yadav2024ties}, DARE-TIES~\cite{yu2024language}, TSV~\cite{tsv}, CART~\cite{choi2024revisiting}, and Iso-C~\cite{marczak2025task}.
This setup isolates the effect of merge space: PGSB only changes coordinate space where LoRA updates are represented before applying downstream merge operator.

\begin{table}[!ht]
    \centering
    \renewcommand\arraystretch{0.9} 
    \caption{Normalized accuracies of merged models on the visual tasks with ViT-B/32.}
    \resizebox{\textwidth}{!}{
    \begin{tabular}{L{2.1cm}L{1.5cm}*{9}{C{1cm}}}
    \toprule
    \textbf{Method} & \textbf{Space} & \textbf{Cars} & \textbf{DTD} & \textbf{EuroSAT} & \textbf{GTSRB} & \textbf{MNIST} & \textbf{RESISC} & \textbf{SUN397} & \textbf{SVHN} & \textbf{Avg} \\
    \midrule
    \textit{Abs. Accuracy} 
        &        & 74.00 & 58.30 & 99.00 & 92.70 & 99.30 & 88.40 & 64.50 & 96.20 & -     \\
    \midrule
    TA  &        & 83.62 & 76.00 & 50.77 & 44.12 & 55.39 & 72.36 & 97.05 & 41.31 & 65.08 \\
    \midrule
    \multirow{4}{*}{TIES} 
        & Baseline   & 82.24 & 73.26 & 51.44 & 37.92 & 56.28 & 69.88 & 96.95 & 43.89 & 63.98 \\
        & KnOTS  & 82.58 & 73.45 & 48.26 & 41.83 & 57.40 & 70.66 & 96.58 & 45.60 & 64.54 \\
        & Core   & 84.29 & 79.74 & 54.10 & 49.67 & 64.50 & 72.72 & 97.44 & 44.29 & 68.34 \\
        & \PGSBcell{PGSB}  & \PGSBcell{83.37} & \PGSBcell{76.37} & \PGSBcell{58.40} & \PGSBcell{44.29} & \PGSBcell{82.18} & \PGSBcell{72.08} & \PGSBcell{95.41} & \PGSBcell{66.17} & \PGSBavg{72.28} \\
    \midrule
    \multirow{4}{*}{DARE-TIES} 
        & Baseline   & 82.35 & 73.26 & 51.44 & 37.92 & 56.26 & 69.96 & 97.01 & 43.81 & 64.00 \\
        & KnOTS  & 82.22 & 75.36 & 51.18 & 40.71 & 56.19 & 71.48 & 97.93 & 39.92 & 64.37 \\
        & Core   & 85.13 & 79.56 & 57.69 & 46.59 & 65.91 & 72.96 & 97.55 & 46.82 & 69.03 \\
        & \PGSBcell{PGSB}  & \PGSBcell{83.45} & \PGSBcell{75.27} & \PGSBcell{58.03} & \PGSBcell{43.01} & \PGSBcell{83.43} & \PGSBcell{71.54} & \PGSBcell{95.13} & \PGSBcell{67.97} & \PGSBavg{72.23} \\
    \midrule
    \multirow{4}{*}{TSV} 
        & Baseline   & 83.40 & 75.45 & 52.49 & 44.99 & 59.50 & 73.26 & 96.33 & 49.31 & 66.84 \\
        & KnOTS  & 81.97 & 75.09 & 51.96 & 41.62 & 53.93 & 71.68 & 97.90 & 40.29 & 64.30 \\
        & Core   & 83.96 & 75.36 & 53.05 & 45.35 & 58.27 & 73.01 & 97.64 & 45.27 & 66.49 \\
        & \PGSBcell{PGSB}  & \PGSBcell{84.04} & \PGSBcell{78.19} & \PGSBcell{54.96} & \PGSBcell{47.32} & \PGSBcell{61.23} & \PGSBcell{74.03} & \PGSBcell{97.59} & \PGSBcell{45.39} & \PGSBavg{67.84} \\
    \midrule
    \multirow{4}{*}{CART}  
        & Baseline   & 83.96 & 77.64 & 52.04 & 48.98 & 56.04 & 73.76 & 97.05 & 43.45 & 66.62 \\
        & KnOTS  & 83.40 & 73.63 & 50.69 & 47.98 & 60.57 & 72.79 & 96.83 & 49.12 & 66.88 \\
        & Core   & 84.04 & 77.00 & 50.77 & 48.30 & 60.12 & 73.13 & 96.46 & 46.17 & 67.00 \\
        & \PGSBcell{PGSB}  & \PGSBcell{83.20} & \PGSBcell{76.00} & \PGSBcell{47.48} & \PGSBcell{49.62} & \PGSBcell{62.89} & \PGSBcell{72.92} & \PGSBcell{96.11} & \PGSBcell{49.56} & \PGSBavg{67.22} \\
    \midrule
    \multirow{4}{*}{$\text{TIES}_\text{+Iso-C}$} 
        & Baseline   & 81.88 & 75.18 & 60.08 & 38.76 & 63.26 & 69.88 & 98.95 & 40.34 & 66.04 \\
        & KnOTS  & 80.77 & 80.56 & 54.88 & 63.95 & 66.38 & 77.26 & 98.81 & 48.95 & 71.45 \\
        & Core   & 84.46 & 84.03 & 57.69 & 75.46 & 68.93 & 81.83 & 99.47 & 50.42 & 75.29 \\
        & \PGSBcell{PGSB}  & \PGSBcell{85.13} & \PGSBcell{84.94} & \PGSBcell{59.63} & \PGSBcell{76.66} & \PGSBcell{73.52} & \PGSBcell{81.90} & \PGSBcell{98.57} & \PGSBcell{51.76} & \PGSBavg{76.51} \\
    \midrule
    \multirow{4}{*}{$\text{DARE-TIES}_\text{+Iso-C}$}
        & Baseline   & 82.14 & 76.18 & 55.85 & 39.71 & 61.00 & 70.05 & 98.71 & 40.47 & 65.51 \\
        & KnOTS  & 76.38 & 81.75 & 45.98 & 65.35 & 68.50 & 78.07 & 97.71 & 46.57 & 70.04 \\
        & Core   & 83.40 & 82.66 & 56.86 & 80.24 & 69.37 & 82.26 & 99.13 & 53.18 & 75.89 \\
        & \PGSBcell{PGSB}  & \PGSBcell{85.30} & \PGSBcell{85.03} & \PGSBcell{59.60} & \PGSBcell{76.79} & \PGSBcell{73.62} & \PGSBcell{81.97} & \PGSBcell{98.73} & \PGSBcell{51.74} & \PGSBavg{76.60} \\
    \midrule
    \multirow{4}{*}{$\text{TSV}_\text{+Iso-C}$} 
        & Baseline   & 78.04 & 80.38 & 61.92 & 66.54 & 65.99 & 79.83 & 98.37 & 45.24 & 72.04 \\
        & KnOTS  & 80.69 & 83.66 & 58.36 & 74.76 & 67.89 & 79.80 & 98.55 & 50.10 & 74.23 \\
        & Core   & 82.81 & 85.03 & 52.45 & 84.25 & 71.01 & 84.32 & 99.01 & 53.55 & 76.55 \\
        & \PGSBcell{PGSB}  & \PGSBcell{84.54} & \PGSBcell{85.40} & \PGSBcell{53.61} & \PGSBcell{84.09} & \PGSBcell{73.36} & \PGSBcell{85.02} & \PGSBcell{98.49} & \PGSBcell{53.75} & \PGSBavg{77.28} \\
    \midrule
    \multirow{4}{*}{$\text{CART}_\text{+Iso-C}$}   
        & Baseline   & 81.59 & 77.83 & 54.4  & 44.76 & 52.81 & 73.06 & 98.91 & 36.88 & 65.03 \\
        & KnOTS  & 82.05 & 79.10 & 55.93 & 55.24 & 56.56 & 76.10 & 99.70 & 41.45 & 68.27 \\
        & Core   & 84.54 & 84.21 & 75.13 & 57.13 & 69.29 & 83.03 & 99.43 & 51.24 & 75.82 \\
        & \PGSBcell{PGSB}  & \PGSBcell{84.38} & \PGSBcell{85.49} & \PGSBcell{52.75} & \PGSBcell{82.17} & \PGSBcell{74.22} & \PGSBcell{83.42} & \PGSBcell{98.45} & \PGSBcell{54.27} & \PGSBavg{76.89} \\
    \midrule
    \multirow{4}{*}{Iso-C} 
        & Baseline   & 80.41 & 82.48 & 53.65 & 74.12 & 69.91 & 79.99 & 98.62 & 49.61 & 73.60 \\
        & KnOTS  & 80.62 & 83.30 & 53.61 & 75.67 & 70.41 & 82.09 & 98.66 & 50.82 & 74.40 \\
        & Core   & 82.95 & 84.94 & 38.76 & 83.15 & 77.39 & 82.09 & 98.18 & 57.47 & 75.62 \\
        & \PGSBcell{PGSB}  & \PGSBcell{84.38} & \PGSBcell{85.40} & \PGSBcell{50.88} & \PGSBcell{82.47} & \PGSBcell{74.02} & \PGSBcell{83.67} & \PGSBcell{98.53} & \PGSBcell{54.67} & \PGSBavg{76.75} \\

    \bottomrule
    \end{tabular}
    } 
    \label{tab:per-task-vitb}
\end{table}

\begin{table}[!ht]
    \centering
    \renewcommand\arraystretch{0.9} 
    \setlength{\tabcolsep}{3mm}
    \caption{Normalized accuracies of merged models on the visual tasks with ViT-L/14.}
    \resizebox{\textwidth}{!}{
    \begin{tabular}{L{2.2cm}L{1.5cm}*{9}{C{1cm}}}
    \toprule
    \textbf{Method} & \textbf{Space} & \textbf{Cars} & \textbf{DTD} & \textbf{EuroSAT} & \textbf{GTSRB} & \textbf{MNIST} & \textbf{RESISC} & \textbf{SUN397} & \textbf{SVHN} & \textbf{Avg} \\
    \midrule
    \textit{Abs. Accuracy} 
        &        & 99.77 & 70.05 & 98.59 & 97.20 & 99.53 & 95.70 & 79.60 & 97.72 & -     \\
    \midrule
    TA  &        & 80.07 & 79.73 & 62.10 & 59.96 & 82.17 & 79.53 & 86.71 & 64.78 & 74.38 \\
    \midrule
    \multirow{4}{*}{TIES} 
        & Baseline   & 79.54 & 78.66 & 63.11 & 60.42 & 84.12 & 78.94 & 87.03 & 69.60 & 75.18 \\
        & KnOTS  & 80.35 & 79.57 & 62.06 & 62.95 & 85.43 & 79.81 & 87.57 & 70.99 & 76.09 \\
        & Core   & 81.91 & 79.88 & 63.07 & 68.95 & 87.48 & 81.77 & 88.50 & 73.76 & 78.17 \\
        & \PGSBcell{PGSB}  & \PGSBcell{83.15} & \PGSBcell{80.18} & \PGSBcell{65.48} & \PGSBcell{69.37} & \PGSBcell{87.85} & \PGSBcell{82.29} & \PGSBcell{87.99} & \PGSBcell{77.52} & \PGSBavg{79.23} \\
    \midrule
    \multirow{4}{*}{DARE-TIES} 
        & Baseline   & 79.56 & 78.66 & 62.96 & 60.39 & 84.11 & 78.95 & 86.99 & 69.61 & 75.15 \\
        & KnOTS  & 80.43 & 79.65 & 62.02 & 63.09 & 85.29 & 79.81 & 87.56 & 71.04 & 76.11 \\
        & Core   & 82.27 & 79.95 & 62.21 & 68.15 & 88.19 & 82.14 & 88.82 & 74.33 & 78.26 \\
        & \PGSBcell{PGSB}  & \PGSBcell{82.34} & \PGSBcell{80.11} & \PGSBcell{65.44} & \PGSBcell{69.04} & \PGSBcell{89.82} & \PGSBcell{82.46} & \PGSBcell{87.87} & \PGSBcell{77.21} & \PGSBavg{79.29} \\
    \midrule
    \multirow{4}{*}{TSV} 
        & Baseline   & 82.10 & 79.88 & 64.05 & 66.79 & 87.31 & 82.09 & 87.56 & 71.67 & 77.68 \\
        & KnOTS  & 80.21 & 79.95 & 64.43 & 60.21 & 83.75 & 79.32 & 86.84 & 65.57 & 75.03 \\
        & Core   & 82.13 & 79.88 & 63.97 & 66.85 & 87.26 & 82.10 & 87.56 & 71.60 & 77.67 \\
        & \PGSBcell{PGSB}  & \PGSBcell{82.50} & \PGSBcell{80.95} & \PGSBcell{64.93} & \PGSBcell{66.88} & \PGSBcell{87.65} & \PGSBcell{83.32} & \PGSBcell{87.63} & \PGSBcell{72.87} & \PGSBavg{78.34} \\
    \midrule
    \multirow{4}{*}{CART} 
        & Baseline   & 91.20 & 89.60 & 76.37 & 81.66 & 74.10 & 93.28 & 86.87 & 66.01 & 82.39 \\
        & KnOTS  & 82.27 & 81.17 & 62.43 & 67.66 & 71.38 & 83.16 & 88.24 & 69.26 & 75.70 \\
        & Core   & 86.02 & 86.33 & 72.39 & 82.82 & 91.03 & 85.93 & 88.46 & 72.67 & 83.21 \\
        & \PGSBcell{PGSB} & \PGSBcell{92.97} & \PGSBcell{88.40} & \PGSBcell{75.10} & \PGSBcell{80.05} & \PGSBcell{90.63} & \PGSBcell{92.90} & \PGSBcell{86.11} & \PGSBcell{66.48} & \PGSBavg{84.08} \\
    \midrule
    \multirow{4}{*}{Iso-C} 
        & Baseline   & 86.96 & 85.42 & 79.94 & 78.40 & 92.19 & 88.21 & 87.22 & 69.09 & 83.43 \\
        & KnOTS  & 88.27 & 88.61 & 77.24 & 82.45 & 87.74 & 89.43 & 88.55 & 68.63 & 83.86 \\
        & Core   & 91.48 & 90.74 & 77.87 & 85.83 & 87.40 & 91.29 & 89.66 & 70.75 & 85.63 \\
        & \PGSBcell{PGSB}  & \PGSBcell{91.45} & \PGSBcell{91.66} & \PGSBcell{77.76} & \PGSBcell{85.94} & \PGSBcell{87.38} & \PGSBcell{91.33} & \PGSBcell{89.69} & \PGSBcell{71.33} & \PGSBavg{85.82} \\
    \bottomrule
    \end{tabular}
    } 
    \label{tab:per-task-vitl}
\end{table}

\begin{table}[!ht]
    \centering
    \renewcommand\arraystretch{0.9} 
    \caption{Normalized accuracies of merged models on the NLI tasks for Llama 3 8B.}
    \resizebox{\textwidth}{!}{
    \begin{tabular}{L{2.5cm}L{1.8cm}*{7}{C{1.3cm}}}
        \toprule
        \textbf{Method} & \textbf{Space} 
        & \textbf{SNLI} & \textbf{MNLI} & \textbf{SICK} & \textbf{QNLI} 
        & \textbf{RTE} & \textbf{SCITAIL} & \textbf{Avg} \\
        \midrule
        \textit{Abs. Accuracy}  
        & -- & 92.50 & 90.31 & 91.58 & 94.49 & 89.86 & 96.52 & -- \\
        \midrule
        TA  
        & -- & 93.58 & 95.28 & 88.07 & 68.76 & 100.00 & 96.73 & 90.40 \\
        \midrule
        
        \multirow{4}{*}{TIES} 
        & Baseline  & 95.17 & 96.21 & 84.20 & 74.21 & 100.00 & 96.73 & 91.09 \\
        & KnOTS & 92.67 & 94.82 & 91.03 & 78.93 & 101.61 & 97.22 & 92.72 \\
        & Core  & 91.94 & 91.12 & 96.68 & 79.27 & 100.81 & 97.37 & 92.86 \\
        & \PGSBcell{PGSB} & \PGSBcell{97.19} & \PGSBcell{97.02} & \PGSBcell{87.35} & \PGSBcell{87.54} & \PGSBcell{95.28} & \PGSBcell{97.97} & \PGSBavg{93.73} \\
        \midrule
        
        \multirow{4}{*}{DARE-TIES}  
        & Baseline  & 94.83 & 96.47 & 83.44 & 74.84 & 98.39 & 96.00 & 90.66 \\
        & KnOTS & 92.95 & 95.53 & 90.70 & 76.58 & 100.00 & 96.98 & 92.12 \\
        & Core  & 92.38 & 93.02 & 93.59 & 84.86 & 100.00 & 97.39 & 93.54 \\
        & \PGSBcell{PGSB} & \PGSBcell{96.97} & \PGSBcell{96.93} & \PGSBcell{88.03} & \PGSBcell{86.20} & \PGSBcell{99.19} & \PGSBcell{97.47} & \PGSBavg{94.13} \\
        \midrule
        
        \multirow{4}{*}{TSV}  
        & Baseline  & 93.95 & 96.05 & 82.26 & 80.00 & 99.19 & 96.88 & 91.39 \\
        & KnOTS & 94.56 & 91.61 & 92.99 & 69.58 & 102.42 & 97.56 & 91.45 \\
        & Core  & 95.86 & 95.71 & 89.27 & 83.94 & 102.42 & 97.86 & 94.18 \\
        & \PGSBcell{PGSB} & \PGSBcell{96.12} & \PGSBcell{96.21} & \PGSBcell{89.43} & \PGSBcell{84.11} & \PGSBcell{102.01} & \PGSBcell{98.02} & \PGSBavg{94.31} \\
        \midrule
        
        \multirow{4}{*}{Iso-C} 
        & Baseline  & 55.00 & 39.04 & 76.54 & 55.90 & 46.77 & 69.25 & 57.08 \\
        & KnOTS & 85.28 & 52.86 & 89.43 & 54.90 & 75.00 & 77.73 & 72.53 \\
        & Core  & 91.54 & 90.10 & 87.87 & 75.85 & 99.19 & 97.42 & 90.33 \\
        & \PGSBcell{PGSB} & \PGSBcell{90.80} & \PGSBcell{90.88} & \PGSBcell{86.69} & \PGSBcell{78.72} & \PGSBcell{101.68} & \PGSBcell{98.52} & \PGSBavg{91.22} \\
        \bottomrule
    \end{tabular}
    }
    \label{tab:per-task-llama}
\end{table}

All hyper-parameters are selected on validation sets, and test labels are never used for model selection.
Implementation details, compute resources, baselines, datasets, and hyperparameter search protocols are provided in Appendix~\ref{app:detail}.

\subsection{Visual Tasks}\label{exp:visual}

Table~\ref{tab:per-task-vitb} reports the normalized accuracies on the visual benchmarks using ViT-B/32~\cite{dosovitskiy2020image}.
Across different merge operators, PGSB consistently provides a stronger merging space than Baseline, KnOTS, and Core Space.
The improvements are especially pronounced for coordinate-sensitive operators such as TIES and DARE-TIES, whose trimming and conflict-resolution steps directly depend on the chosen coordinate system.
We further evaluate PGSB with a stronger visual backbone, ViT-L/14, and the results are shown in Table~\ref{tab:per-task-vitl}.
PGSB again achieves the best average performance across the evaluated settings, demonstrating that the proposed pretrained-compatible merging space remains effective across different visual backbones.

\subsection{Natural Language Inference}

Table~\ref{tab:per-task-llama} reports the normalized accuracies on the NLI benchmarks using Llama 3 8B~\cite{llama3modelcard}.
PGSB achieves the best average performance across all evaluated merge operators when compared with Baseline, KnOTS~\cite{stoica2024knots}, and Core Space~\cite{panariello2026core}.
These results indicate that the proposed merging space is not specific to visual encoders, but also transfers effectively to large language models.
They further support the generality of representing LoRA updates in pretrained-compatible coordinates before applying downstream merge operators.

\subsection{Commensense Reasoning}

Table~\ref{tab:cr} reports the normalized accuracies on the Commonsense Reasoning benchmarks using Llama 3 8B~\cite{llama3modelcard}.
We use TIES as the merge operator and compare PGSB against Baseline, KnOTS~\cite{stoica2024knots}, and Core Space~\cite{panariello2026core}.
PGSB achieves the best average performance of 97.58\%, outperforming all competing merging spaces.
These results further demonstrate that PGSB effectively preserves reasoning capabilities during model merging.

\begin{table}[!ht]
    \centering
    \renewcommand\arraystretch{0.9} 
    \caption{Normalized accuracies of merged models on the Commonsense Reasoning tasks for Llama 3 8B with TIES merging.}
    \resizebox{\textwidth}{!}{
    \begin{tabular}{L{2.2cm}*{9}{C{1.3cm}}}
        \toprule
        \textbf{Space} 
        & \textbf{ARC-c} & \textbf{ARC-e} & \textbf{BoolQ} & \textbf{HellaS.} 
        & \textbf{OBQA} & \textbf{PIQA} & \textbf{SIQA} & \textbf{WinoG.} & \textbf{Avg} \\
        \midrule
        \textit{Abs. Accuracy} & 80.38 & 92.89 & 89.51 & 96.28 & 86.2 & 90.32 & 80.6 & 88.08 & -- \\
        \midrule
         Baseline  & 99.68 & 99.23 & 97.56 & 92.8 & 96.29 & 96.75 & 98.1 & 97.13 & 97.19 \\
         KnOTS & 102.34 & 100.3 & 98.14 & 90.11 & 96.98 & 97.11 & 97.02 & 95.96 & 97.25 \\
         Core  & 100.42 & 99.18 & 95.66 & 92.47 & 96.52 & 95.3 & 96.26 & 95.16 & 96.37 \\
         \PGSBcell{PGSB} & \PGSBcell{100.21} & \PGSBcell{99.55} & \PGSBcell{97.66} & \PGSBcell{93.72} & \PGSBcell{97.68} & \PGSBcell{97.11} & \PGSBcell{97.46} & \PGSBcell{97.22} & \PGSBavg{97.58} \\
        
        \bottomrule
    \end{tabular}
    }
    \label{tab:cr}
\end{table}

\subsection{Ablation Study}

We study the necessity of the two stages of PGSB through two ablation variants.
Both variants start from the task-shared support pair $\mathbf{U}_L,\mathbf{U}_R$ extracted by TSE.
\emph{TSE-only} removes PCR and directly merges task updates in the original support coordinates:
\begin{equation}
    \mathbf{D}_t=\mathbf{U}_L^\top \Delta \mathbf{W}_t \mathbf{U}_R,
    \qquad
    \mathbf{D}^\star=\mathcal{M}(\mathbf{D}_1,\dots,\mathbf{D}_T).
\end{equation}
The merged update is then reconstructed as
$\Delta\mathbf{W}^\star=\mathbf{U}_L\mathbf{D}^\star\mathbf{U}_R^\top$.
\emph{No-weighted} keeps PCR but removes the singular-value weighting.
Instead of decomposing the weighted pretrained coupling
$\mathbf{T}=\mathbf{\Sigma}_L(\mathbf{U}_L^\top\mathbf{W}_0\mathbf{U}_R)\mathbf{\Sigma}_R$,
it directly decomposes the unweighted coupling
$\mathbf{G}=\mathbf{U}_L^\top\mathbf{W}_0\mathbf{U}_R
=\widetilde{\mathbf{O}}_L\mathbf{\Sigma}_G\widetilde{\mathbf{O}}_R^\top$,
and constructs
$\mathbf{Q}_L=\mathbf{U}_L\widetilde{\mathbf{O}}_L$ and
$\mathbf{Q}_R=\mathbf{U}_R\widetilde{\mathbf{O}}_R$.

Table~\ref{tab:ablation} shows that \emph{TSE-only} reaches an average score of $68.46$, outperforming the Baseline under TIES merging.
This indicates that TSE provides an effective task-shared support pair for representing LoRA updates.
Full PGSB further improves the score to $72.28$, showing that PCR provides additional gains beyond support construction by refining the coordinates inside the same support.
In contrast, \emph{No-weighted} drops to $62.86$, suggesting that directly decomposing $\mathbf{G}$ is insufficient for obtaining merge-friendly coordinates.
Weighting $\mathbf{G}$ by both $\mathbf{\Sigma}_L$ and $\mathbf{\Sigma}_R$ is therefore important for emphasizing task-shared directions, reducing harmful coordinate mixing, and improving coordinate-sensitive merging.

\begin{table}[!ht]
\centering
\caption{Ablations on the visual tasks with TIES merging. Each entry reports normalized accuracy on the corresponding dataset. Baseline directly applies TIES to the LoRA updates without any other operations.}
\resizebox{\textwidth}{!}{
    \begin{tabular}{L{3cm}*{9}{C{1cm}}}
    \toprule
    \textbf{Variant} & \textbf{Cars} & \textbf{DTD} & \textbf{EuroSAT} & \textbf{GTSRB} & \textbf{MNIST} & \textbf{RESISC} & \textbf{SUN397} & \textbf{SVHN} & \textbf{Avg} \\
    \midrule
    PGSB & 83.37 & 76.37 & 58.40 & 44.29 & 82.18 & 72.08 & 95.41 & 66.17 & 72.28 \\
    \midrule
    Baseline & 82.24 & 73.26 & 51.44 & 37.92 & 56.28 & 69.88 & 96.95 & 43.89 & 63.98 \\
    TSE-only & 84.88 & 77.37 & 54.55 & 46.81 & 65.46 & 72.31 & 96.69 & 49.64 & 68.46 \\
    No-weighted  & 81.76 & 74.63 & 51.37 & 36.91 & 53.58 & 69.54 & 98.04 & 37.06 & 62.86 \\
    \bottomrule
    \end{tabular}
}
\label{tab:ablation}
\end{table}

\subsection{Extensibility Study}

We next examine whether PCR is specific to the support pair constructed by PGSB, or whether it can serve as a general coordinate refinement module for existing alignment-based merge spaces.
To this end, we keep the shared spaces produced by KnOTS~\cite{stoica2024knots} and Core Space~\cite{panariello2026core} unchanged, and apply PCR on top of them to obtain \emph{KnOTS + PCR} and \emph{Core Space + PCR}.
These hybrid variants preserve the original support spaces of prior methods, but replace their inherited coordinates with pretrained-compatible coordinates refined by PCR.
Implementation details for applying PCR to prior merge spaces are provided in Appendix~\ref{app:pcr_prior_spaces}.

\begin{table}[!ht]
\centering
\caption{Extensibility experiments on the visual tasks with TIES merging. Each entry reports normalized accuracy on the corresponding dataset.}
\resizebox{\textwidth}{!}{
    \begin{tabular}{L{3cm}*{9}{C{1cm}}}
    \toprule
    \textbf{Variant} & \textbf{Cars} & \textbf{DTD} & \textbf{EuroSAT} & \textbf{GTSRB} & \textbf{MNIST} & \textbf{RESISC} & \textbf{SUN397} & \textbf{SVHN} & \textbf{Avg} \\
    \midrule
    PGSB             & 83.37 & 76.37 & 58.40 & 44.29 & 82.18 & 72.08 & 95.41 & 66.17 & 72.28 \\
    \midrule
    KnOTS            & 82.58 & 73.45 & 48.26 & 41.83 & 57.40 & 70.66 & 96.58 & 45.60 & 64.54 \\
    KnOTS + PCR      & 83.00 & 72.26 & 56.12 & 37.34 & 70.88 & 70.41 & 96.66 & 61.28 & 68.49 \\
    Core Space       & 84.29 & 79.74 & 54.10 & 49.67 & 64.50 & 72.72 & 97.44 & 44.29 & 68.34 \\
    Core Space + PCR & 83.79 & 77.00 & 50.24 & 47.75 & 77.04 & 68.07 & 96.19 & 57.93 & 69.75 \\
    \bottomrule
    \end{tabular}
}
\label{tab:extensibility}
\end{table}

Table~\ref{tab:extensibility} shows that PCR consistently improves prior alignment spaces.
Specifically, \emph{KnOTS + PCR} increases the average score from $64.54$ to $68.49$, while \emph{Core Space + PCR} improves it from $68.34$ to $69.75$.
These results indicate that pretrained-compatible coordinate refinement is not tied to PGSB's own support construction, but can also enhance existing merge spaces.
Nevertheless, neither hybrid reaches the performance of full PGSB, which achieves an average score of $72.28$.
This suggests that PCR is broadly extensible, while the strongest results are obtained when it is combined with the task-shared support constructed by PGSB.

\subsection{Pretrained Transfer Preservation Study}

We further evaluate whether the merged model preserves the transfer ability inherited from the pretrained backbone.
The motivation is that LoRA updates are learned as task-specific offsets to the same pretrained weight $\mathbf{W}_0$.
After merging, the final model is obtained by adding the merged update back to $\mathbf{W}_0$.
Therefore, an effective merge space should not only combine knowledge from the source tasks, but also avoid unnecessarily distorting the pretrained representation, so that the model can still generalize to tasks outside the merge set.

Concretely, we use the same eight-task visual merging setting as in Sec.~\ref{exp:visual}.
For each merge space, including Baseline, KnOTS, Core Space, and PGSB, we merge the same eight task-specific LoRA updates using TIES~\cite{yadav2024ties}.
The resulting merged update is then added to the CLIP ViT-B/32 pretrained weight~\cite{radford2021learning,dosovitskiy2020image}.
Without any additional training, fine-tuning, or adaptation, we directly evaluate the merged models on seven additional visual datasets that are not used for training the source LoRA adapters or for constructing the merged update.
These datasets cover diverse visual domains, including generic object recognition, fine-grained animal classification, food recognition, flower recognition, and natural image classification, with representative datasets such as CIFAR100~\cite{krizhevsky2009learning}, Caltech101~\cite{li2004cal}, and Food101~\cite{Bossard2014Food101M}.

Table~\ref{tab:preservation} reports normalized accuracies, using the original pretrained model as the reference.
PGSB achieves the highest average normalized accuracy of $98.38$, remaining closest to the pretrained model among all merged models.
Compared with Baseline, KnOTS, and Core Space, PGSB better preserves performance on tasks outside the merge set.
These results show that PGSB not only improves multi-task merging on the source tasks, but also better maintains the general representation ability of the pretrained model after merging.

\begin{table}[!ht]
\centering
\caption{Normalized accuracies of merged models on additional visual tasks after TIES merging. \emph{Abs. Accuracy} reports the absolute accuracy of the pretrained model, which is used as the normalization reference for each visual task.
}
\resizebox{\textwidth}{!}{
    \begin{tabular}{lcccccccc}
    \toprule
    \textbf{Space} & \textbf{CIFAR10} & \textbf{CIFAR100} & \textbf{Pets} & \textbf{Flowers102} & \textbf{Food101} & \textbf{Caltech101} & \textbf{Cats vs Dogs} & \textbf{Avg} \\
    \midrule
    \textit{Abs. Accuracy} & 88.86 & 64.35 & 86.24 & 65.98 & 84.43 & 79.90 & 99.62 & 81.34 \\
    \midrule
    Baseline        & 97.60 & 78.57 & 95.63 & 91.77 & 97.95 & 96.72 & 99.99 & 94.03 \\
    KnOTS           & 81.87 & 30.27 & 85.74 & 61.90 & 89.92 & 85.47 & 99.94 & 76.44 \\
    Core Space      & 99.07 & 90.58 & 97.84 & 97.01 & 98.65 & 99.50 & 99.96 & 97.52 \\
    \PGSBcell{PGSB} & \PGSBcell{99.32} & \PGSBcell{94.65} & \PGSBcell{98.92} & \PGSBcell{97.71} & \PGSBcell{99.04} & \PGSBcell{98.97} & \PGSBcell{100.00} & \PGSBavg{98.38} \\
    \bottomrule
    \end{tabular}
}
\label{tab:preservation}
\end{table}

\subsection{Rank Robustness Study}

We further investigate the robustness of PGSB under diverse LoRA rank configurations on the eight visual tasks using ViT-B/32.
Specifically, we consider three settings: (1) homogeneous LoRA ranks of $r \in \{4,64,256\}$, covering a wide range of adaptation capacities; (2) heterogeneous ranks, where Cars, DTD, EuroSAT, and GTSRB use $r=16$, while MNIST, RESISC45, SUN397, and SVHN use $r=64$; and (3) adaptive ranks, where all eight task adapters are trained using AdaLoRA~\cite{zhang2023adalora} with a target rank of $r=16$.
For all settings, we apply TIES~\cite{yadav2024ties} as the merge operator and compare PGSB against Baseline, KnOTS~\cite{stoica2024knots}, and Core Space~\cite{panariello2026core}.

Table~\ref{tab:rank_robustness} reports the normalized accuracies under these configurations.
PGSB achieves the highest average performance in all five settings, demonstrating its effectiveness across homogeneous, heterogeneous, and adaptively allocated ranks.
These results demonstrate that PGSB does not rely on a fixed or uniform LoRA rank and remains effective across diverse rank configurations.

\begin{table}[!ht]
    \centering
    \renewcommand\arraystretch{0.9}
    \caption{Normalized accuracies of merged models under different LoRA rank configurations on the eight visual tasks using ViT-B/32 with TIES merging. }
    \resizebox{\textwidth}{!}{
    \begin{tabular}{L{1.8cm}L{2.3cm}*{9}{C{1.3cm}}}
        \toprule
        \textbf{Setting} & \textbf{Space}
        & \textbf{Cars} & \textbf{DTD} & \textbf{EuroSAT}
        & \textbf{GTSRB} & \textbf{MNIST} & \textbf{RESISC45}
        & \textbf{SUN397} & \textbf{SVHN} & \textbf{Avg} \\
        \midrule

        \multirow{5}{*}{$r=4$}
        & \textit{Abs. Accuracy}
        & 68.57 & 68.83 & 98.19 & 96.65 & 99.49 & 91.60 & 71.34 & 96.05 & -- \\
        & Baseline
        & 89.59 & 72.26 & 60.62 & 63.59 & 82.75 & 80.64 & 93.68 & 79.71 & 77.85 \\
        & KnOTS
        & 92.74 & 75.50 & 61.60 & 68.37 & 82.13 & 78.06 & 93.56 & 76.39 & 78.54 \\
        & Core
        & 92.38 & 76.66 & 66.09 & 66.59 & 79.32 & 77.54 & 92.98 & 72.31 & 77.98 \\
        & \PGSBcell{PGSB}
        & \PGSBcell{92.81} & \PGSBcell{75.81} & \PGSBcell{68.58}
        & \PGSBcell{65.45} & \PGSBcell{86.83} & \PGSBcell{81.13}
        & \PGSBcell{93.33} & \PGSBcell{81.66} & \PGSBavg{80.70} \\
        \midrule

        \multirow{5}{*}{$r=64$}
        & \textit{Abs. Accuracy}
        & 74.63 & 75.32 & 98.56 & 97.87 & 99.50 & 93.38 & 73.48 & 96.64 & -- \\
        & Baseline
        & 82.94 & 69.28 & 69.00 & 73.73 & 90.85 & 81.54 & 90.30 & 89.01 & 80.83 \\
        & KnOTS
        & 86.36 & 70.62 & 68.96 & 76.24 & 90.35 & 79.69 & 89.56 & 88.17 & 81.24 \\
        & Core
        & 87.58 & 78.67 & 80.72 & 79.04 & 86.63 & 82.95 & 91.05 & 79.46 & 83.26 \\
        & \PGSBcell{PGSB}
        & \PGSBcell{86.33} & \PGSBcell{82.49} & \PGSBcell{67.91}
        & \PGSBcell{83.08} & \PGSBcell{89.20} & \PGSBcell{83.78}
        & \PGSBcell{91.26} & \PGSBcell{83.36} & \PGSBavg{83.43} \\
        \midrule

        \multirow{5}{*}{$r=256$}
        & \textit{Abs. Accuracy}
        & 77.81 & 76.70 & 99.04 & 98.39 & 99.65 & 94.54 & 73.97 & 96.87 & -- \\
        & Baseline
        & 79.19 & 64.36 & 74.20 & 71.38 & 90.69 & 77.25 & 89.03 & 85.62 & 78.96 \\
        & KnOTS
        & 81.95 & 63.31 & 66.57 & 65.05 & 90.25 & 75.65 & 89.07 & 86.51 & 77.30 \\
        & Core
        & 83.44 & 68.86 & 74.91 & 70.81 & 88.96 & 79.13 & 90.51 & 80.09 & 79.59 \\
        & \PGSBcell{PGSB}
        & \PGSBcell{82.32} & \PGSBcell{72.19} & \PGSBcell{72.70}
        & \PGSBcell{71.94} & \PGSBcell{88.56} & \PGSBcell{80.42}
        & \PGSBcell{90.44} & \PGSBcell{80.51} & \PGSBavg{79.89} \\
        \midrule

        \multirow{5}{*}{Heterogeneous}
        & \textit{Abs. Accuracy}
        & 74.00 & 58.30 & 99.00 & 92.70 & 99.50 & 93.38 & 73.48 & 96.64 & -- \\
        & Baseline
        & 86.31 & 74.72 & 69.47 & 40.66 & 89.00 & 80.16 & 89.41 & 78.69 & 76.05 \\
        & KnOTS
        & 86.97 & 76.00 & 64.05 & 41.92 & 87.15 & 78.63 & 89.23 & 79.85 & 75.47 \\
        & Core
        & 85.71 & 77.73 & 65.73 & 43.43 & 88.96 & 83.05 & 90.81 & 76.46 & 76.49 \\
        & \PGSBcell{PGSB}
        & \PGSBcell{87.49} & \PGSBcell{76.18} & \PGSBcell{69.32}
        & \PGSBcell{44.97} & \PGSBcell{91.86} & \PGSBcell{85.30}
        & \PGSBcell{89.92} & \PGSBcell{82.23} & \PGSBavg{78.41} \\
        \midrule

        \multirow{5}{*}{$\text{AdaLoRA}_\text{r=16}$}
        & \textit{Abs. Accuracy}
        & 73.84 & 58.56 & 97.70 & 95.34 & 99.27 & 89.79 & 69.29 & 95.49 & -- \\
        & Baseline
        & 94.56 & 88.83 & 58.95 & 70.87 & 81.28 & 82.46 & 95.05 & 80.10 & 81.51 \\
        & KnOTS
        & 95.31 & 92.33 & 65.66 & 71.60 & 82.35 & 82.36 & 95.03 & 78.70 & 82.92 \\
        & Core
        & 95.43 & 91.46 & 71.19 & 80.62 & 80.19 & 87.45 & 95.80 & 78.73 & 85.11 \\
        & \PGSBcell{PGSB}
        & \PGSBcell{96.74} & \PGSBcell{91.35} & \PGSBcell{73.01}
        & \PGSBcell{78.61} & \PGSBcell{84.57} & \PGSBcell{88.99}
        & \PGSBcell{96.15} & \PGSBcell{78.84} & \PGSBavg{86.03} \\

        \bottomrule
    \end{tabular}
    }
    \label{tab:rank_robustness}
\end{table}

\section{Conclusion}

In this paper, we studied LoRA model merging as a scalable route toward building unified multi-task models from parameter-efficient adaptations.
We identified two key challenges that distinguish LoRA merging from conventional merging of fully fine-tuned models: extracting shared task directions from diverse low-rank updates, and representing these directions in coordinates compatible with the pretrained weight geometry.
To address these challenges, we proposed PGSB, a two-stage framework consisting of Task-shared Subspace Extraction and Pretrained-coupled Coordinate Refinement.
PGSB first constructs a shared subspace from concatenated LoRA updates, and then refines the merge coordinates using the geometric structure of the pretrained weights.
This yields a pretrained-compatible basis that better supports coordinate-sensitive merging and improves multi-task adaptation.
Extensive experiments on vision and language benchmarks demonstrate that PGSB consistently outperforms strong baselines.
Further analyses confirm the necessity of both shared support construction and pretrained-guided coordinate refinement, show that PGSB better preserves the transfer ability of the pretrained model, and demonstrate the robustness of PGSB across diverse rank configurations.

\bibliography{main}
\bibliographystyle{plain}


\newpage

\appendix
\section{Prior Alignment-Based Methods}
\label{app:prior_alignment_methods}

Let the LoRA update for task $t$ be
\begin{equation}
    \mathbf{\Delta W}_t = \mathbf{B}_t \mathbf{A}_t \in \mathbb{R}^{d_{\mathrm{out}} \times d_{\mathrm{in}}},
    \qquad t=1,\dots,T,
\end{equation}
where $\mathbf{B}_t \in \mathbb{R}^{d_{\mathrm{out}} \times r}$ and $\mathbf{A}_t \in \mathbb{R}^{r \times d_{\mathrm{in}}}$ are the low-rank adaptation factors.
Existing alignment-based LoRA merging methods can be viewed as attempts to construct a common representation space for the task updates before applying a merge operator.
Below, we summarize two representative methods and discuss their relation to the two questions considered in PGSB: how to extract shared task directions, and how to choose pretrained-compatible merge coordinates.

\subsection{KnOTS.}
KnOTS aligns LoRA updates by reconstructing the full update matrices and extracting a shared basis from their concatenation.
For a given layer, it computes
\begin{equation}
    [\mathbf{\Delta W}_1,\mathbf{\Delta W}_2,\dots,\mathbf{\Delta W}_T]
    =
    \mathbf{U} \mathbf{\Sigma}
    \begin{bmatrix}
        \mathbf{V}^{(1)}, \mathbf{V}^{(2)}, \cdots , \mathbf{V}^{(T)}
    \end{bmatrix}^{\!\top}.
\end{equation}
Here, $\mathbf{U}$ provides a common left basis shared by all task updates, $\Sigma$ contains the corresponding singular values, and each $\mathbf{V}^{(t)}$ gives the task-specific coordinates under this shared basis.
A merge operator is then applied to $\{\mathbf{V}^{(t)}\}_{t=1}^T$ to obtain $\mathbf{V}^{(\mathrm{merge})}$, and the merged update is reconstructed as
\begin{equation}
    \mathbf{\Delta W}^{(\mathrm{merge})}
    =
    \mathbf{U} \mathbf{\Sigma} \bigl(\mathbf{V}^{(\mathrm{merge})}\bigr)^\top .
\end{equation}

This formulation shows that KnOTS addresses part of the first question: it extracts a shared set of directions from concatenated task updates and performs merging in the resulting aligned space.
However, the alignment is essentially one-sided.
The shared structure is made explicit through the common left basis $U$, while the right-side structure remains absorbed in the task-specific coordinate matrices $V^{(t)}$.
Thus, KnOTS does not explicitly construct a shared left-right support pair for LoRA updates.
Moreover, its merge coordinates are inherited directly from the SVD of task updates and are not further refined using the pretrained weight geometry.
Finally, because KnOTS performs SVD on concatenated full update matrices, its preprocessing cost grows with the ambient weight dimension, which weakens the efficiency advantage of LoRA.

\subsection{Core Space.}
Core Space avoids full-space SVD by constructing shared reference spaces directly from the low-rank LoRA factors.
Specifically, it computes
\begin{equation}
    [\mathbf{B}_1,\mathbf{B}_2,\dots,\mathbf{B}_T]
    =
    \mathbf{U}_B^{\mathrm{ref}} \mathbf{\Sigma}_B \mathbf{V}_B^\top,
\end{equation}
and
\begin{equation}
    \begin{bmatrix}
        \mathbf{A}_1\\ \mathbf{A}_2\\ \vdots\\ \mathbf{A}_T
    \end{bmatrix}
    =
    \mathbf{U}_A \mathbf{\Sigma}_A \bigl(\mathbf{V}_A^{\mathrm{ref}}\bigr)^\top.
\end{equation}
Here, $\mathbf{U}_B^{\mathrm{ref}}$ spans a shared left reference space for the $\mathbf{B}_t$ factors, while $\mathbf{V}_A^{\mathrm{ref}}$ spans a shared right reference space for the $\mathbf{A}_t$ factors.
Each task update is represented by a compact core matrix
\begin{equation}
    \mathbf{M}_t
    =
    \bigl(\mathbf{U}_B^{\mathrm{ref}\top} \mathbf{B}_t\bigr)
    \bigl(\mathbf{A}_t \mathbf{V}_A^{\mathrm{ref}}\bigr),
\end{equation}
and merging is performed on $\{\mathbf{M}_t\}_{t=1}^T$.
Denoting the merged core matrix by $\mathbf{M}^{(\mathrm{merge})}$, the final merged update is reconstructed as
\begin{equation}
    \mathbf{\Delta W}^{(\mathrm{merge})}
    =
    \mathbf{U}_B^{\mathrm{ref}} \mathbf{M}^{(\mathrm{merge})} \mathbf{V}_A^{\mathrm{ref}\top}.
\end{equation}

Compared with KnOTS, Core Space preserves the low-rank regime and provides an efficient way to represent LoRA updates in a shared compact space.
In this sense, it also addresses part of the first question by introducing shared left and right reference spaces.
However, these two spaces are constructed independently from the $B_t$ and $A_t$ factors, rather than from the aggregate structure of the reconstructed task updates.
Consequently, Core Space does not explicitly characterize the relative strength of shared directions in the full update space.
More importantly, as with KnOTS, the merge coordinates are directly induced by the shared-space construction itself and are not coupled with the common pretrained weight $\mathbf{W}_0$.
Therefore, Core Space improves the efficiency of LoRA merging, but it does not explicitly address the second question of how to choose pretrained-compatible coordinates within the shared supports.

Overall, both KnOTS and Core Space demonstrate the importance of aligning LoRA updates before merging.
KnOTS emphasizes shared directions through full-update concatenation, while Core Space emphasizes efficient low-rank representation.
PGSB builds on this alignment perspective, but separates the problem into two stages: first extracting task-shared left and right support subspaces with singular-value-based directional strength, and then refining the resulting basis using the geometry of the pretrained weight.

\subsection{Applying PCR to Prior Methods}
\label{app:pcr_prior_spaces}

We describe how PCR is applied to prior merge spaces in the extensibility study.
The key idea is simple: we keep the support construction of each prior method unchanged, and only apply PCR to rotate the coordinates inside the obtained merge space before applying the downstream merge operator.

\paragraph{KnOTS + PCR.}
For KnOTS, we keep its original one-sided merge form.
After truncation, KnOTS represents each task update as
\begin{equation}
    \Delta\mathbf{W}_t
    \approx
    \mathbf{U}_k\mathbf{\Sigma}_k
    (\mathbf{V}^{(t)}_k)^\top .
\end{equation}
Thus, the shared KnOTS basis is $\mathbf{U}_k\mathbf{\Sigma}_k$, and the merge operator is applied to the task-specific coordinate blocks $(\mathbf{V}^{(t)}_k)^\top$.

To apply PCR without changing this one-sided structure, we refine only the coordinates inside the KnOTS shared basis.
We form the pretrained-coupled matrix
\begin{equation}
    \mathbf{T}_{\mathrm{K}}
    =
    (\mathbf{U}_k\mathbf{\Sigma}_k)^\top \mathbf{W}_0
    =
    \mathbf{\Sigma}_k\mathbf{U}_k^\top\mathbf{W}_0,
\end{equation}
and compute its compact SVD
\begin{equation}
    \mathbf{T}_{\mathrm{K}}
    =
    \mathbf{O}_{\mathrm{K}}
    \mathbf{\Gamma}_{\mathrm{K}}
    \mathbf{R}_{\mathrm{K}}^\top .
\end{equation}
The refined KnOTS basis is
\begin{equation}
    \widehat{\mathbf{U}}_{\mathrm{K}}
    =
    \mathbf{U}_k\mathbf{\Sigma}_k\mathbf{O}_{\mathrm{K}},
\end{equation}
and each task coordinate is correspondingly rotated as
\begin{equation}
    \widehat{\mathbf{V}}^{(t)\top}_{\mathrm{K}}
    =
    \mathbf{O}_{\mathrm{K}}^\top
    (\mathbf{V}^{(t)}_k)^\top .
\end{equation}
This transformation preserves each task update before merging since
\begin{equation}
    \mathbf{U}_k\mathbf{\Sigma}_k
    (\mathbf{V}^{(t)}_k)^\top
    =
    \widehat{\mathbf{U}}_{\mathrm{K}}
    \widehat{\mathbf{V}}^{(t)\top}_{\mathrm{K}} .
\end{equation}
We then apply the downstream merge operator to the refined coordinate blocks:
\begin{equation}
    \widehat{\mathbf{V}}_{\mathrm{K}}^{\mathrm{merge}\top}
    =
    \mathcal{M}
    \left(
    \widehat{\mathbf{V}}^{(1)\top}_{\mathrm{K}},
    \dots,
    \widehat{\mathbf{V}}^{(T)\top}_{\mathrm{K}}
    \right),
\end{equation}
and reconstruct the merged update as
\begin{equation}
    \Delta\mathbf{W}_{\mathrm{K}}^{\mathrm{merge}}
    =
    \widehat{\mathbf{U}}_{\mathrm{K}}
    \widehat{\mathbf{V}}_{\mathrm{K}}^{\mathrm{merge}\top}.
\end{equation}

\paragraph{Core Space + PCR.}
Core Space constructs shared reference spaces from the stacked LoRA factors:
\begin{equation}
    [\mathbf{B}_1,\mathbf{B}_2,\dots,\mathbf{B}_T]
    =
    \mathbf{U}_{B}^{\mathrm{ref}}
    \mathbf{\Sigma}_{B}
    \mathbf{V}_{B}^{\top},
    \qquad
    \begin{bmatrix}
    \mathbf{A}_1\\
    \mathbf{A}_2\\
    \vdots\\
    \mathbf{A}_T
    \end{bmatrix}
    =
    \mathbf{U}_{A}
    \mathbf{\Sigma}_{A}
    (\mathbf{V}_{A}^{\mathrm{ref}})^\top .
\end{equation}
Therefore, its original merge space is given by the left-right support pair
\begin{equation}
    \mathbf{U}_{B}^{\mathrm{ref}},
    \qquad
    \mathbf{V}_{A}^{\mathrm{ref}},
\end{equation}
with strength spectra $\mathbf{\Sigma}_{B}$ and $\mathbf{\Sigma}_{A}$.
PCR keeps these supports fixed and forms
\begin{equation}
    \mathbf{T}_{\mathrm{C}}
    =
    \mathbf{\Sigma}_{B}
    \left(
    (\mathbf{U}_{B}^{\mathrm{ref}})^\top
    \mathbf{W}_0
    \mathbf{V}_{A}^{\mathrm{ref}}
    \right)
    \mathbf{\Sigma}_{A}.
\end{equation}
We compute
\begin{equation}
    \mathbf{T}_{\mathrm{C}}
    =
    \mathbf{O}_{\mathrm{C}}^{L}
    \mathbf{\Gamma}_{\mathrm{C}}
    (\mathbf{O}_{\mathrm{C}}^{R})^\top ,
\end{equation}
and obtain the PCR-refined Core Space bases:
\begin{equation}
    \widehat{\mathbf{U}}_{\mathrm{C}}
    =
    \mathbf{U}_{B}^{\mathrm{ref}}\mathbf{O}_{\mathrm{C}}^{L},
    \qquad
    \widehat{\mathbf{V}}_{\mathrm{C}}
    =
    \mathbf{V}_{A}^{\mathrm{ref}}\mathbf{O}_{\mathrm{C}}^{R}.
\end{equation}
Each task update is represented as
\begin{equation}
    \mathbf{C}^{\mathrm{C}}_t
    =
    \widehat{\mathbf{U}}_{\mathrm{C}}^\top
    \Delta\mathbf{W}_t
    \widehat{\mathbf{V}}_{\mathrm{C}}
    =
    \widehat{\mathbf{U}}_{\mathrm{C}}^\top
    \mathbf{B}_t\mathbf{A}_t
    \widehat{\mathbf{V}}_{\mathrm{C}} .
\end{equation}
The merged coordinate matrix is
\begin{equation}
    \mathbf{C}_{\mathrm{C}}^{\mathrm{merge}}
    =
    \mathcal{M}
    (
    \mathbf{C}^{\mathrm{C}}_1,\dots,\mathbf{C}^{\mathrm{C}}_T
    ),
\end{equation}
and the final merged update is reconstructed as
\begin{equation}
    \Delta\mathbf{W}_{\mathrm{C}}^{\mathrm{merge}}
    =
    \widehat{\mathbf{U}}_{\mathrm{C}}
    \mathbf{C}_{\mathrm{C}}^{\mathrm{merge}}
    \widehat{\mathbf{V}}_{\mathrm{C}}^\top .
\end{equation}

In both cases, PCR does not change how the prior method constructs its merge space.
It only replaces the original coordinates inside that space with pretrained-coupled coordinates, after which the same merge operator and reconstruction rule are applied.

\section{Equivalent Low-Rank Construction of Shared Supports}
\label{app:support_construction}

In this section, we show that the task-shared support pair used in TSE can be obtained without explicitly performing SVD on the concatenated full update matrices.
For one LoRA-adapted weight matrix, define the full-update concatenations as
\begin{equation}
    \mathbf{X}:=
    [\Delta \mathbf{W}_1,\Delta \mathbf{W}_2,\dots,\Delta \mathbf{W}_T]
    =
    \mathbf{U}_L \mathbf{\Sigma}_L \mathbf{V}_L^\top,
\end{equation}
and
\begin{equation}
    \mathbf{Y}:=
    [\Delta \mathbf{W}_1^\top,\Delta \mathbf{W}_2^\top,\dots,\Delta \mathbf{W}_T^\top]
    =
    \mathbf{U}_R \mathbf{\Sigma}_R \mathbf{V}_R^\top.
\end{equation}
The full-update formulation of TSE uses $\mathbf{U}_L$ and $\mathbf{U}_R$ as the shared left and right supports, while $\mathbf{\Sigma}_L$ and $\mathbf{\Sigma}_R$ characterize the aggregate strength of the corresponding shared directions.

For each task $t$, let
\begin{equation}
    \Delta \mathbf{W}_t
    =
    \mathbf{U}_t \mathbf{\Sigma}_t \mathbf{V}_t^\top
\end{equation}
be its compact SVD.
We construct two low-rank concatenations:
\begin{equation}
    \mathbf{L}:=
    [\mathbf{U}_1\mathbf{\Sigma}_1,
     \mathbf{U}_2\mathbf{\Sigma}_2,
     \dots,
     \mathbf{U}_T\mathbf{\Sigma}_T],
    \qquad
    \mathbf{R}:=
    [\mathbf{V}_1\mathbf{\Sigma}_1,
     \mathbf{V}_2\mathbf{\Sigma}_2,
     \dots,
     \mathbf{V}_T\mathbf{\Sigma}_T].
\end{equation}
We now show that the SVDs of $\mathbf{L}$ and $\mathbf{R}$ recover the same shared supports and singular values as the full-update construction.

For the left support, since
\begin{equation}
    \Delta \mathbf{W}_t
    =
    \mathbf{U}_t \mathbf{\Sigma}_t \mathbf{V}_t^\top,
\end{equation}
the full concatenation can be written as
\begin{equation}
    \mathbf{X}
    =
    \mathbf{L}\,
    \operatorname{blkdiag}
    (\mathbf{V}_1^\top,\mathbf{V}_2^\top,\dots,\mathbf{V}_T^\top).
\end{equation}
Because each $\mathbf{V}_t$ has orthonormal columns, the block-diagonal matrix above has orthonormal rows. Therefore,
\begin{equation}
    \mathbf{X}\mathbf{X}^\top
    =
    \mathbf{L}\mathbf{L}^\top.
\end{equation}
Thus, $\mathbf{X}$ and $\mathbf{L}$ have the same left Gram matrix, which implies that they share the same nonzero singular values and the same left singular subspace. Consequently, the shared left support $\mathbf{U}_L$ and the strength spectrum $\mathbf{\Sigma}_L$ can be recovered equivalently from the SVD of $\mathbf{L}$.

The same argument applies to the right support. Since
\begin{equation}
    \mathbf{Y}
    =
    \mathbf{R}\,
    \operatorname{blkdiag}
    (\mathbf{U}_1^\top,\mathbf{U}_2^\top,\dots,\mathbf{U}_T^\top),
\end{equation}
and each $\mathbf{U}_t$ has orthonormal columns, we have
\begin{equation}
    \mathbf{Y}\mathbf{Y}^\top
    =
    \mathbf{R}\mathbf{R}^\top.
\end{equation}
Therefore, $\mathbf{Y}$ and $\mathbf{R}$ also share the same nonzero singular values and the same left singular subspace. Hence, $\mathbf{U}_R$ and $\mathbf{\Sigma}_R$ can be recovered equivalently from the SVD of $\mathbf{R}$.

This proves that the low-rank construction based on $\mathbf{L}$ and $\mathbf{R}$ yields the same task-shared left and right support subspaces as the full-update TSE formulation, while avoiding explicit SVD on the concatenated full update matrices.
The equivalence holds up to the standard sign and rotation ambiguities of SVD within degenerate singular subspaces.

\section{Computational Complexity Analysis}
\label{app:complexity}

We analyze the preprocessing complexity of PGSB for constructing the merge space of one LoRA-adapted layer before applying a downstream merge operator.
Following prior analyses, we assume a square weight matrix of size $n\times n$, $T$ task-specific LoRA updates, and LoRA rank $r$ with $Tr\ll n$.
The complexity of model merging operators, KnOTS, and Core Space follows the analysis in the Core Space paper~\cite{panariello2026core}.

\paragraph{PGSB Complexity.}
The direct formulation of PGSB first reconstructs each task update
\begin{equation}
    \Delta \mathbf{W}_t=\mathbf{B}_t\mathbf{A}_t,
    \qquad t=1,\dots,T,
\end{equation}
which costs $\mathcal{O}(n^2Tr)$ in total.
It then constructs the left and right task-shared supports by applying SVD to the concatenated full updates
\begin{equation}
    [\Delta\mathbf{W}_1,\dots,\Delta\mathbf{W}_T],
    \qquad
    [\Delta\mathbf{W}_1^\top,\dots,\Delta\mathbf{W}_T^\top].
\end{equation}
Each concatenated matrix has size $n\times nT$, and the full-space SVD cost is $O(n^3T^2)$ under the same convention as KnOTS.
Since PGSB performs this operation on both the update matrix and its transpose, the asymptotic order remains $\mathcal{O}(n^3T^2)$.

After obtaining the support matrices $\mathbf{U}_L,\mathbf{U}_R$ and strength spectra $\mathbf{\Sigma}_L,\mathbf{\Sigma}_R$, PGSB performs pretrained-coupled coordinate refinement.
Let $k\leq Tr$ be the dimension of the shared support.
Computing
\begin{equation}
    \mathbf{G}=\mathbf{U}_L^\top\mathbf{W}_0\mathbf{U}_R
\end{equation}
costs $\mathcal{O}(n^2k)$, which is upper-bounded by $\mathcal{O}(n^2Tr)$.
The SVD of the resulting $k\times k$ matrix
\begin{equation}
    \mathbf{T}=\mathbf{\Sigma}_L\mathbf{G}\mathbf{\Sigma}_R
\end{equation}
costs $\mathcal{O}(k^3)$, which is upper-bounded by $\mathcal{O}(T^3r^3)$.
Therefore, the total preprocessing complexity of the direct PGSB formulation is $\mathcal{O}(n^3T^2+n^2Tr+T^3r^3)$.

\paragraph{PGSB (Efficient) Complexity.}
The efficient implementation avoids SVD on concatenated full-size update matrices.
For each task update, let
\begin{equation}
    \Delta\mathbf{W}_t
    =
    \mathbf{U}_t\mathbf{\Sigma}_t\mathbf{V}_t^\top
\end{equation}
be its compact SVD.
Instead of factorizing $[\Delta\mathbf{W}_1,\dots,\Delta\mathbf{W}_T]$ and
$[\Delta\mathbf{W}_1^\top,\dots,\Delta\mathbf{W}_T^\top]$, PGSB constructs the compact matrices
\begin{equation}
    \mathbf{L}
    =
    [\mathbf{U}_1\mathbf{\Sigma}_1,\dots,\mathbf{U}_T\mathbf{\Sigma}_T],
    \qquad
    \mathbf{R}
    =
    [\mathbf{V}_1\mathbf{\Sigma}_1,\dots,\mathbf{V}_T\mathbf{\Sigma}_T],
\end{equation}
whose widths are at most $Tr$.
SVD on these low-rank concatenations costs $\mathcal{O}(n(Tr)^2+(Tr)^3)$,
which is no larger than the subsequent pretrained-coupled refinement cost under $Tr\ll n$.
The refinement step is the same as in PGSB: computing
$\mathbf{G}=\mathbf{U}_L^\top\mathbf{W}_0\mathbf{U}_R$ costs $\mathcal{O}(n^2Tr)$, and decomposing the $Tr\times Tr$ matrix $\mathbf{T}$ costs $\mathcal{O}(T^3r^3)$.
Thus, the efficient implementation has total preprocessing complexity $\mathcal{O}(n^2Tr+T^3r^3)$.

This matches the asymptotic order of Core Space while preserving the two-sided task-shared support construction and the pretrained-coupled coordinate refinement of PGSB.

\section{Algorithm table of PGSB}\label{app:algo}

The full procedure of PGSB is summarized in Algorithm ~\ref{alg:PGSB}.

\begin{algorithm}[!ht]
\caption{PGSB for LoRA Model Merging}
\label{alg:PGSB}
\begin{algorithmic}[1]
\Require Pretrained weights $\{\mathbf{W}_0^\ell\}_{\ell=1}^L$, task updates $\{\Delta \mathbf{W}_t^\ell\}_{t=1,\ell=1}^{T,L}$, merge rule $\mathcal{M}$
\Ensure Merged model $\mathbf{W}_{\mathrm{merge}}$

\For{each layer $\ell$}
    \State Compute compact SVDs $\Delta \mathbf{W}_t^\ell=\mathbf{U}_t^\ell \mathbf{\Sigma}_t^\ell \mathbf{V}_t^{\ell\top}$ for all $t$
    \State Compute $[\mathbf{U}_1^\ell\mathbf{\Sigma}_1^\ell,\dots,\mathbf{U}_T^\ell\mathbf{\Sigma}_T^\ell]=\mathbf{U}_L^\ell\mathbf{\Sigma}_L^\ell\widetilde{\mathbf{V}}_L^{\ell\top}$ and $[\mathbf{V}_1^\ell\mathbf{\Sigma}_1^\ell,\dots,\mathbf{V}_T^\ell\mathbf{\Sigma}_T^\ell]=\mathbf{U}_R^\ell\mathbf{\Sigma}_R^\ell\widetilde{\mathbf{V}}_R^{\ell\top}$
    \State Form $\mathbf{T}^\ell=\mathbf{\Sigma}_L^\ell(\mathbf{U}_L^{\ell\top}\mathbf{W}_0^\ell\mathbf{U}_R^\ell)\mathbf{\Sigma}_R^\ell$
    \State Decompose $\mathbf{T}^\ell=\mathbf{O}_L^\ell\mathbf{\Sigma}_T^\ell\mathbf{O}_R^{\ell\top}$
    \State Set $\mathbf{Q}_L^\ell=\mathbf{U}_L^\ell\mathbf{O}_L^\ell$ and $\mathbf{Q}_R^\ell=\mathbf{U}_R^\ell\mathbf{O}_R^\ell$
    \State Compute $\mathbf{C}_t^\ell=\mathbf{Q}_L^{\ell\top}\Delta \mathbf{W}_t^\ell\mathbf{Q}_R^\ell$ for all $t$, then merge $\mathbf{C}_*^\ell=\mathcal{M}(\{\mathbf{C}_t^\ell\}_{t=1}^T)$
    \State Reconstruct $\Delta \mathbf{W}_*^\ell=\mathbf{Q}_L^\ell\mathbf{C}_*^\ell\mathbf{Q}_R^{\ell\top}$
\EndFor
\State \Return $\mathbf{W}_{\mathrm{merge}}=\mathbf{W}_0+\{\Delta \mathbf{W}_*^\ell\}_{\ell=1}^L$
\end{algorithmic}
\end{algorithm}

\section{Experiment Details}\label{app:detail}

All the experiments were performed on NVIDIA RTX 3090. 
Our implementation builds on the KnOTS codebase and uses the released LoRA checkpoints from~\cite{stoica2024knots}.
For the original training and adaptation procedures, we refer readers to~\cite{stoica2024knots}.

\subsection{Baselines}

\paragraph{Merging spaces.}
We compare our proposed PGSB with three alternative merging spaces, which are all plugins. 
\begin{itemize}
    \item \textbf{Baseline} performs merging on the  updates $\Delta \mathbf{W}^{(t)} = \mathbf{B}^{(t)}\mathbf{A}^{(t)}$ directly. 
    \item \textbf{KnOTS}~\cite{stoica2024knots} performs merging in the alignment space constructed from the concatenated updates. 
    \item \textbf{Core Space}~\cite{panariello2026core} performs merging in the shared low-rank basis induced by the stacked LoRA factors. 
    \item \textbf{PGSB} performs merging in pretrained-guided shared basis proposed in Eq.\ref{eq:PGSB}.
\end{itemize}

\paragraph{Merging methods.}
We evaluate each merging space using representative merging methods. 
\begin{itemize}
    \item \textbf{Task Arithmetic (TA)}~\cite{ilharco2022editing} performs scaled summation of task vectors. 
    \item \textbf{TIES}~\cite{yadav2024ties} trims small-magnitude parameters and resolves sign conflicts before averaging. 
    \item \textbf{DARE-TIES}~\cite{yu2024language} further applies random dropping and rescaling before TIES merging.
    \item \textbf{TSV}~\cite{tsv} concatenates low-rank approximations of task updates and orthogonalizes them across tasks. 
    \item \textbf{CART}~\cite{choi2024revisiting} first centers task vectors by subtracting their average and then performs merging on the resulting low-rank approximations. 
    \item \textbf{Iso-C}~\cite{marczak2025task} flattens the singular value spectrum of a merged model and can be combined with other merging methods. We denote such variants by ``+Iso-C''. 
\end{itemize}
Since TA is linear, its results are equivalent across different merging spaces after exact reconstruction, whereas the choice of merging space mainly affects the performance of nonlinear methods.

\subsection{Datasets}

\paragraph{Vision tasks.}

We evaluate vision model merging on eight image classification datasets: Cars196~\cite{cars}, DTD~\cite{dtd}, EuroSAT~\cite{eurosat}, GTSRB~\cite{gtsrb}, MNIST~\cite{MNIST}, NWPU-RESISC45~\cite{cheng2017remote}, SUN397~\cite{sun397}, and SVHN~\cite{svhn}.

To evaluate generalization beyond the original merging tasks, we further use seven additional datasets: CIFAR10, CIFAR100~\cite{krizhevsky2009learning}, Oxford-IIIT Pets~\cite{parkhi2012pets}, Oxford Flowers102~\cite{nilsback_automated_2008}, Food101~\cite{Bossard2014Food101M}, Caltech101~\cite{li2004cal}, and Cats vs Dogs~\cite{elson2007catsvsdogs}.

\paragraph{Language tasks.}

We evaluate language model merging on six NLI datasets: SNLI~\cite{bowman2015large}, MNLI~\cite{williams2017broad}, SICK~\cite{marelli2014sick}, QNLI~\cite{wang2018glue}, RTE~\cite{wang2018glue}, and SciTail~\cite{scitail}.

We also evaluate language model merging on eight Commonsense Reasoning datasets: ARC-Challenge and ARC-Easy~\cite{clark2018thinkarce}, BoolQ~\cite{clark2019boolq}, HellaSwag~\cite{zellers2019hellaswag}, OpenBookQA~\cite{mihaylov2018canobqa}, PIQA~\cite{bisk2020piqa}, Social IQa~\cite{sap2019socialiqa}, and WinoGrande~\cite{sakaguchi2021winogrande}.

\paragraph{Dataset usage.}

All datasets are public benchmark datasets and are used only for research and benchmarking purposes, following their original licenses or terms of use.

\subsection{Hyperparameter Search Strategy}

To ensure fair comparison, we follow prior works~\cite{stoica2024knots,panariello2026core} and apply the same validation-based hyperparameter search protocol to all merging spaces and merge operators.
PGSB only changes the coordinate space in which task updates are represented and does not introduce a new downstream merge operator; therefore, the hyper-parameters of each merge operator are selected in the same way across all compared spaces.
No test-set labels are used during hyperparameter selection.

For methods with an overall scaling coefficient, we search $\alpha$ from $0.1$ with step size $0.1$ until validation performance stops improving.
For TIES~\cite{yadav2024ties} and DARE-TIES~\cite{yu2024language}, we additionally search the top-$K$ trimming parameter from $10$ with step size $10$.
For DARE-TIES, the pruning factor $p$ is searched from $0.1$ with step size $0.1$.
For CART~\cite{choi2024revisiting}, we follow the original setting and search the pruning rank over $\{0.04,0.08,0.16,0.32\}$.
CART also introduces an additional scaling coefficient $\lambda$:
\begin{equation}
\mathbf{W}_{\mathrm{merge}}=\mathbf{W}_0+\alpha\left(\Delta\mathbf{W}_{\mathrm{avg}}+\lambda\sum_{t=1}^{T}\bar{\tau}_t\right),
\end{equation}
where $\Delta\mathbf{W}_{\mathrm{avg}}$ denotes the average update and $\bar{\tau}_t$ denotes the centered task vector for task $t$.
We search $\lambda$ using the same validation strategy as $\alpha$.
For variants combined with Iso-C~\cite{marczak2025task}, we first apply the corresponding merge operator with selected hyper-parameters and then apply Iso-C post-processing.

\subsection{Computational Efficiency and Memory Analysis}
\label{app:efficiency}

We further evaluate the practical computational efficiency of PGSB in terms of merging time, peak GPU memory consumption, and hyperparameter search overhead.
Although PGSB introduces an additional pretrained-coupled coordinate refinement (PCR) stage, its implementation processes the pretrained weights layer by layer and releases intermediate tensors immediately, avoiding the need to maintain an additional full copy of $W_0$ on the GPU.
Moreover, the shared bases and SVD results are independent of the downstream merging hyperparameters.
Therefore, these quantities can be computed once and cached for reuse across subsequent merges and hyperparameter search trials.

\paragraph{Merging Time and Memory Consumption.}
Table~\ref{tab:merge_efficiency} compares the merging time and peak GPU memory consumption of different merging spaces on ViT-B/32, ViT-L/14, and Llama 3 8B.
We report both uncached and cached results for PGSB, where the uncached setting includes the construction of the shared bases and PCR, while the cached setting reuses the precomputed quantities.

Without caching, PGSB incurs additional computational overhead compared with Core Space due to PCR.
However, caching substantially reduces the merging time to 0.84s, 1.39s, and 12.43s for ViT-B/32, ViT-L/14, and Llama 3 8B, respectively.
Consequently, cached PGSB is faster than Core Space on both vision backbones and achieves comparable efficiency on Llama 3 8B.
Compared with KnOTS, cached PGSB achieves a $6.27\times$--$30.66\times$ speedup while reducing peak GPU memory consumption by 62.9\%--85.5\%.
These results indicate that the additional coordinate refinement does not introduce prohibitive computational or memory overhead when intermediate computations are reused.

\begin{table}[!ht]
    \centering
    \renewcommand\arraystretch{0.9}
    \caption{Merging time and peak GPU memory consumption of different merging spaces. PGSB (uncached) includes shared basis construction and PCR, while PGSB (cached) reuses precomputed intermediate results.}
    \resizebox{\textwidth}{!}{
    \begin{tabular}{L{2cm}L{3.5cm}C{2.5cm}C{4cm}}
        \toprule
        \textbf{Model} & \textbf{Merge Space}
        & \textbf{Merge Time (s)}
        & \textbf{Peak GPU Memory (MiB)} \\
        \midrule

        \multirow{5}{*}{ViT-B/32}
        & Baseline & 0.73 & 103.5 \\
        & KnOTS & 5.27 & 90.6 \\
        & Core Space & 1.55 & 20.5 \\
        & PGSB (uncached) & 5.76 & 60.1 \\
        & PGSB (cached) & 0.84 & 33.6 \\
        \midrule

        \multirow{5}{*}{ViT-L/14}
        & Baseline & 2.50 & 184.0 \\
        & KnOTS & 19.46 & 153.1 \\
        & Core Space & 3.01 & 28.3 \\
        & PGSB (uncached) & 13.48 & 87.6 \\
        & PGSB (cached) & 1.39 & 44.1 \\
        \midrule

        \multirow{5}{*}{Llama 3 8B}
        & Baseline & 26.51 & 2272.0 \\
        & KnOTS & 381.11 & 1750.5 \\
        & Core Space & 11.43 & 273.6 \\
        & PGSB (uncached) & 39.94 & 502.6 \\
        & PGSB (cached) & 12.43 & 254.7 \\

        \bottomrule
    \end{tabular}
    }
    \label{tab:merge_efficiency}
\end{table}

\paragraph{Hyperparameter Search Efficiency.}
The dimensions of the shared bases in PGSB are determined by the numerical ranks of the aggregated task supports at each layer, rather than by the downstream hyperparameter search.
Thus, the shared subspace construction and PCR only need to be performed once for each set of task adapters.
Subsequent hyperparameter configurations can directly reuse the cached representations without repeating the SVD computations.

To assess the practical impact, we measure the total hyperparameter search time on ViT-B/32 using TIES under the same linear-search protocol.
As shown in Table~\ref{tab:search_efficiency}, PGSB requires 727s, compared with 806s for Baseline, 794s for KnOTS, and 1187s for Core Space.
PGSB therefore achieves the lowest total search time among the evaluated merging spaces, demonstrating that its additional coordinate refinement does not impose an extra validation-search burden in practice.

\begin{table}[!ht]
    \centering
    \renewcommand\arraystretch{0.9}
    \caption{Total hyperparameter search time on ViT-B/32 using TIES under the same linear-search protocol.}
    \begin{tabular}{L{2.8cm}C{2.8cm}}
        \toprule
        \textbf{Merge Space} & \textbf{Search Time (s)} \\
        \midrule
        Baseline & 806 \\
        KnOTS & 794 \\
        Core Space & 1187 \\
        PGSB & \textbf{727} \\
        \bottomrule
    \end{tabular}
    \label{tab:search_efficiency}
\end{table}

Overall, layer-wise processing and caching substantially reduce the computational and memory overhead of PGSB.
The resulting implementation achieves competitive merging efficiency and low validation-search cost while retaining the superior merging performance demonstrated in the main experiments.

\section{Newer Backbones Experiments}
\label{app:siglip}

To further evaluate the generalization ability of PGSB beyond CLIP-based visual encoders, we conduct additional experiments on the more recent SigLIP backbone~\cite{zhai2023sigmoid}.
We merge LoRA adapters trained on the same eight visual classification tasks as in the main experiments, using TIES~\cite{yadav2024ties} as the merge operator.

Table~\ref{tab:siglip_merging} reports the normalized accuracies of the merged models.
PGSB achieves the highest average performance of 91.27, outperforming Baseline (89.13), KnOTS~\cite{stoica2024knots} (88.86), and Core Space~\cite{panariello2026core} (90.91).
These results indicate that the effectiveness of PGSB is not limited to the CLIP backbones evaluated in the main experiments, but also extends to newer vision encoders such as SigLIP.

\begin{table}[!ht]
    \centering
    \renewcommand\arraystretch{0.9}
    \caption{Normalized accuracies of merged models on the eight visual tasks using SigLIP-B/16 with TIES merging.}
    \resizebox{\textwidth}{!}{
    \begin{tabular}{L{2.3cm}*{9}{C{1.3cm}}}
        \toprule
        \textbf{Space}
        & \textbf{Cars} & \textbf{DTD} & \textbf{EuroSAT}
        & \textbf{GTSRB} & \textbf{MNIST} & \textbf{RESISC45}
        & \textbf{SUN397} & \textbf{SVHN} & \textbf{Avg} \\
        \midrule
        \textit{Abs. Accuracy}
        & 94.13 & 81.81 & 98.78 & 97.89 & 99.53 & 94.21 & 76.93 & 97.27 & -- \\
        \midrule
        Baseline
        & 98.00 & 89.73 & 58.42 & 90.88 & 97.83 & 84.99 & 94.82 & 98.38 & 89.13 \\
        KnOTS
        & 98.61 & 89.79 & 56.81 & 92.85 & 98.34 & 82.70 & 94.99 & 96.80 & 88.86 \\
        Core
        & 99.05 & 91.35 & 65.73 & 95.85 & 98.34 & 84.85 & 95.46 & 96.61 & 90.91 \\
        \PGSBcell{PGSB}
        & \PGSBcell{98.57} & \PGSBcell{95.19} & \PGSBcell{63.33}
        & \PGSBcell{92.78} & \PGSBcell{98.51} & \PGSBcell{85.93}
        & \PGSBcell{97.84} & \PGSBcell{97.99} & \PGSBavg{91.27} \\
        \bottomrule
    \end{tabular}
    }
    \label{tab:siglip_merging}
\end{table}

\section{Heterogeneous Tasks Merging Experiments}
\label{app:heterogeneous_tasks}

We further investigate the ability of PGSB to merge LoRA adapters trained on heterogeneous language tasks using Llama 3 8B~\cite{llama3modelcard}.
Unlike the homogeneous task configurations evaluated in the main experiments, we jointly merge three commonsense reasoning tasks (OBQA, PIQA, and WinoGrande) and three natural language inference tasks (QNLI, RTE, and SciTail).
We adopt TIES~\cite{yadav2024ties} as the merge operator and compare PGSB with Baseline, KnOTS~\cite{stoica2024knots}, and Core Space~\cite{panariello2026core}.

Table~\ref{tab:heterogeneous_tasks} reports the normalized accuracies on the six tasks.
PGSB achieves the highest average score of 97.06, surpassing Baseline (96.64), KnOTS (96.12), and Core Space (96.73).
These results demonstrate that PGSB remains effective when merging adapters from different task families, supporting its applicability to heterogeneous language task collections.

\begin{table}[!ht]
    \centering
    \renewcommand\arraystretch{0.9}
    \caption{Normalized accuracies of merged models on heterogeneous language tasks using Llama 3 8B with TIES merging.}
    \resizebox{\textwidth}{!}{
    \begin{tabular}{L{2.5cm}*{7}{C{1.5cm}}}
        \toprule
        \textbf{Space}
        & \textbf{OBQA} & \textbf{PIQA} & \textbf{WinoG.}
        & \textbf{QNLI} & \textbf{RTE} & \textbf{SciTail} & \textbf{Avg} \\
        \midrule
        \textit{Abs. Accuracy}
        & 86.20 & 90.32 & 88.08 & 94.49 & 89.86 & 96.52 & -- \\
        \midrule
        Baseline
        & 93.27 & 93.55 & 99.55 & 96.97 & 98.38 & 98.10 & 96.64 \\
        KnOTS
        & 94.43 & 94.58 & 100.63 & 90.81 & 97.98 & 98.29 & 96.12 \\
        Core
        & 100.93 & 95.54 & 98.92 & 88.51 & 99.19 & 97.27 & 96.73 \\
        \PGSBcell{PGSB}
        & \PGSBcell{96.98} & \PGSBcell{96.08} & \PGSBcell{98.65}
        & \PGSBcell{95.08} & \PGSBcell{97.98} & \PGSBcell{97.56}
        & \PGSBavg{97.06} \\
        \bottomrule
    \end{tabular}
    }
    \label{tab:heterogeneous_tasks}
\end{table}

\section{PCR Effectiveness Analysis}
\label{app:pcr_factors}

We further investigate how different factors affect the effectiveness of Pretrained-coupled Coordinate Refinement (PCR).
To isolate its contribution, we compare the full PGSB framework against TSE-only, which performs merging directly in the task-shared support coordinates without PCR.
We measure the improvement introduced by PCR as
$\Delta_{\mathrm{PCR}} = \mathrm{PGSB} - \mathrm{TSE\text{-}only}$,
where the difference is computed using the average normalized accuracy across the evaluated tasks.

Specifically, we consider four factors: task similarity, number of adapters, LoRA rank, and merge operator.
For task similarity, we measure the average pairwise cosine similarity between task updates and select the three most similar and three least similar four-task groups.
For the number of adapters, we randomly sample three task subsets for each of $T=2$ and $T=4$, while $T=8$ includes all eight tasks.
We additionally vary the LoRA rank over $\{4,16,64,256\}$ and evaluate four representative merge operators: TIES, DARE-TIES, TSV, and CART.

Table~\ref{tab:pcr_factors} summarizes the performance improvements introduced by PCR under these settings.
PCR yields substantially larger gains for highly similar tasks (+4.40) than for less similar tasks (+0.84), suggesting that pretrained-guided coordinate refinement is particularly effective when task updates exhibit coherent shared structures.
PCR also consistently improves performance across different numbers of adapters and LoRA ranks, although the gains are non-monotonic, indicating that its effectiveness depends on task composition rather than adapter count or rank alone.

Regarding merge operators, PCR achieves the largest improvements with TIES (+3.82) and DARE-TIES (+5.15).
Both methods rely on coordinate-wise pruning and sign conflict resolution, making their merging behavior sensitive to the choice of basis.
In contrast, the gains are smaller for TSV (+0.07) and CART (+0.99), which incorporate their own subspace transformations.
Overall, these results indicate that PCR is most beneficial when task updates share coherent geometric structures and the downstream merge operator is sensitive to the coordinate representation.

\begin{table}[!ht]
    \centering
    \renewcommand\arraystretch{0.95}
    \caption{Effect of different factors on the performance improvement introduced by PCR. $\Delta_{\mathrm{PCR}}$ denotes the difference in average normalized accuracy between PGSB and TSE-only. Positive values indicate improvements from PCR.}
    \resizebox{\textwidth}{!}{
    \begin{tabular}{llc}
        \toprule
        \textbf{Factor} & \textbf{Setting} & $\boldsymbol{\Delta_{\mathrm{PCR}}}$ \\
        \midrule
        Task similarity
        & High / Low
        & +4.40 / +0.84 \\
        \midrule
        Number of adapters
        & $T=2/4/8$
        & +1.30 / +5.68 / +3.82 \\
        \midrule
        LoRA rank
        & $r=4/16/64/256$
        & +2.12 / +3.82 / +3.11 / +1.63 \\
        \midrule
        Merge operator
        & TIES / DARE-TIES / TSV / CART
        & +3.82 / +5.15 / +0.07 / +0.99 \\
        \bottomrule
    \end{tabular}}
    \label{tab:pcr_factors}
\end{table}

\newpage
\section*{NeurIPS Paper Checklist}

\begin{enumerate}

\item {\bf Claims}
    \item[] Question: Do the main claims made in the abstract and introduction accurately reflect the paper's contributions and scope?
    \item[] Answer: \answerYes{} 
    \item[] Justification: The main claims made in the abstract and introduction accurately reflect our paper’s contributions and scope.

\item {\bf Limitations}
    \item[] Question: Does the paper discuss the limitations of the work performed by the authors?
    \item[] Answer: \answerYes{}
    \item[] Justification: We discuss this in the conclusion section.
    \begin{itemize}
        \item The answer \answerNA{} means that the paper has no limitation while the answer \answerNo{} means that the paper has limitations, but those are not discussed in the paper. 
        \item The authors are encouraged to create a separate ``Limitations'' section in their paper.
        \item The paper should point out any strong assumptions and how robust the results are to violations of these assumptions (e.g., independence assumptions, noiseless settings, model well-specification, asymptotic approximations only holding locally). The authors should reflect on how these assumptions might be violated in practice and what the implications would be.
        \item The authors should reflect on the scope of the claims made, e.g., if the approach was only tested on a few datasets or with a few runs. In general, empirical results often depend on implicit assumptions, which should be articulated.
        \item The authors should reflect on the factors that influence the performance of the approach. For example, a facial recognition algorithm may perform poorly when image resolution is low or images are taken in low lighting. Or a speech-to-text system might not be used reliably to provide closed captions for online lectures because it fails to handle technical jargon.
        \item The authors should discuss the computational efficiency of the proposed algorithms and how they scale with dataset size.
        \item If applicable, the authors should discuss possible limitations of their approach to address problems of privacy and fairness.
        \item While the authors might fear that complete honesty about limitations might be used by reviewers as grounds for rejection, a worse outcome might be that reviewers discover limitations that aren't acknowledged in the paper. The authors should use their best judgment and recognize that individual actions in favor of transparency play an important role in developing norms that preserve the integrity of the community. Reviewers will be specifically instructed to not penalize honesty concerning limitations.
    \end{itemize}

\item {\bf Theory assumptions and proofs}
    \item[] Question: For each theoretical result, does the paper provide the full set of assumptions and a complete (and correct) proof?
    \item[] Answer: \answerYes{}
    \item[] Justification: All the assumptions proposed in this paper have been proven.

    \item {\bf Experimental result reproducibility}
    \item[] Question: Does the paper fully disclose all the information needed to reproduce the main experimental results of the paper to the extent that it affects the main claims and/or conclusions of the paper (regardless of whether the code and data are provided or not)?
    \item[] Answer: \answerYes{}
    \item[] Justification: We provide all implementation details.

\item {\bf Open access to data and code}
    \item[] Question: Does the paper provide open access to the data and code, with sufficient instructions to faithfully reproduce the main experimental results, as described in supplemental material?
    \item[] Answer: \answerNo{}
    \item[] Justification: We will provide codes in the near future.

\item {\bf Experimental setting/details}
    \item[] Question: Does the paper specify all the training and test details (e.g., data splits, hyperparameters, how they were chosen, type of optimizer) necessary to understand the results?
    \item[] Answer: \answerYes{}
    \item[] Justification: We provide all the experimental details.

\item {\bf Experiment statistical significance}
    \item[] Question: Does the paper report error bars suitably and correctly defined or other appropriate information about the statistical significance of the experiments?
    \item[] Answer: \answerNo{}
    \item[] Justification: We use official evaluation protocols provided by the corresponding benchmarks.

\item {\bf Experiments compute resources}
    \item[] Question: For each experiment, does the paper provide sufficient information on the computer resources (type of compute workers, memory, time of execution) needed to reproduce the experiments?
    \item[] Answer: \answerYes{}
    \item[] Justification: We provide all the implementation details.
    
\item {\bf Code of ethics}
    \item[] Question: Does the research conducted in the paper conform, in every respect, with the NeurIPS Code of Ethics \url{https://neurips.cc/public/EthicsGuidelines}?
    \item[] Answer: \answerYes{}
    \item[] Justification: We have checked our work.

\item {\bf Broader impacts}
    \item[] Question: Does the paper discuss both potential positive societal impacts and negative societal impacts of the work performed?
    \item[] Answer: \answerYes{}
    \item[] Justification: We discuss this in the conclusion and introduction sections.
    
\item {\bf Safeguards}
    \item[] Question: Does the paper describe safeguards that have been put in place for responsible release of data or models that have a high risk for misuse (e.g., pre-trained language models, image generators, or scraped datasets)?
    \item[] Answer: \answerNA{}
    \item[] Justification: The paper poses no such risks.

\item {\bf Licenses for existing assets}
    \item[] Question: Are the creators or original owners of assets (e.g., code, data, models), used in the paper, properly credited and are the license and terms of use explicitly mentioned and properly respected?
    \item[] Answer: \answerYes{}
    \item[] Justification: All are properly referred to.

\item {\bf New assets}
    \item[] Question: Are new assets introduced in the paper well documented and is the documentation provided alongside the assets?
    \item[] Answer: \answerNA{}
    \item[] Justification: The paper does not release new assets.

\item {\bf Crowdsourcing and research with human subjects}
    \item[] Question: For crowdsourcing experiments and research with human subjects, does the paper include the full text of instructions given to participants and screenshots, if applicable, as well as details about compensation (if any)? 
    \item[] Answer: \answerNA{}
    \item[] Justification: No such experiments.

\item {\bf Institutional review board (IRB) approvals or equivalent for research with human subjects}
    \item[] Question: Does the paper describe potential risks incurred by study participants, whether such risks were disclosed to the subjects, and whether Institutional Review Board (IRB) approvals (or an equivalent approval/review based on the requirements of your country or institution) were obtained?
    \item[] Answer: \answerNA{}
    \item[] Justification: No such experiments.

\item {\bf Declaration of LLM usage}
    \item[] Question: Does the paper describe the usage of LLMs if it is an important, original, or non-standard component of the core methods in this research? Note that if the LLM is used only for writing, editing, or formatting purposes and does \emph{not} impact the core methodology, scientific rigor, or originality of the research, declaration is not required.
    \item[] Answer: \answerNA{}
    \item[] Justification: Not use LLM at all.

\end{enumerate}

\end{document}